\documentclass[journal]{IEEEtran}

\usepackage{amsmath}
\usepackage{amssymb}
\usepackage{graphicx}
\usepackage{algorithm} 
\usepackage{algorithmic}
\usepackage{xspace}
\usepackage{url}
\usepackage{morefloats}
\usepackage{cite}
\usepackage{color}
\usepackage[usenames,dvipsnames]{xcolor}
\usepackage[tight,footnotesize]{subfigure}

\newcommand{\bea}{\begin{eqnarray}}
\newcommand{\eea}{\end{eqnarray}}

\begin{document}

\title{Search-Space Characterization for \\ Real-time Heuristic Search}

\author{Daniel Huntley and Vadim Bulitko \\
Department of Computing Science, University of Alberta, \\ Edmonton, Alberta, T6G 2E8, Canada \\ {\small\tt \{dhuntley $\mid$ bulitko\}@ualberta.ca}}

\maketitle

\begin{abstract}
Recent real-time heuristic search algorithms have demonstrated outstanding performance in video-game pathfinding.  However, their applications have been thus far limited to that domain.  We proceed with the aim of facilitating wider applications of real-time search by fostering a greater understanding of the performance of recent algorithms.  We first introduce eight algorithm-independent complexity measures for search spaces and correlate their values with algorithm performance.  The complexity measures are statistically shown to be significant predictors of algorithm performance across a set of commercial video-game maps.  We then extend this analysis to a wider variety of search spaces in the first application of database-driven real-time search to domains outside of video-game pathfinding. In doing so, we gain insight into algorithm performance and possible enhancement as well as into search space complexity.
\end{abstract}


\section{Introduction}
\label{chap:intro}

Heuristic search is a mainstay of artificial intelligence research.  A demand for quickly generated solutions to search problems gave rise to the sub-field of \emph{real-time heuristic search}.  Real-time heuristic search algorithms make decisions in constant time, independent of the number of states in the problem being solved.

Recent database-driven real-time heuristic search algorithms have demonstrated excellent performance in video-game pathfinding~\cite{hcdps-ajc}.  However, despite being formulated for general search, most of these algorithms have not been applied to a broader selection of problems.  In preliminary experimentation, we found that many of the algorithms yielded mixed or poor results in other search spaces, such as combinatorial search puzzles.  Motivated by this mixed performance, we seek to establish a way of empirically characterizing search spaces based on their suitability for different real-time heuristic search algorithms.  This would assist algorithm selection, and provide insight for the development of future algorithms.

In this section we discuss the goals of our research.  We then describe our specific contributions, and outline the layout of the remainder of this paper.


\subsection{Motivation}

There are three goals motivating our research. First, we seek a greater understanding of where database-driven real-time heuristic search is and is not effective.  Second, we wish to demonstrate that this knowledge can facilitate algorithm selection and parameterization.  Finally, we want to provide a way for other researchers to characterize benchmark search problems for the development of future real-time heuristic search algorithms. We address these goals through an analysis of real-time heuristic search performance, correlating it to search space features.  


\subsection{Contributions}

This paper makes the following contributions.  First, we present a set of complexity measures which are useful for characterizing search space complexity as it pertains to the performance of modern real-time search algorithms.  Some of these complexity measures are original to this work, and some have been adapted from the literature.

We then empirically link the values of the collected complexity measures to the performance of modern database-driven real-time search algorithms.  We begin with an examination of algorithm performance in pathfinding on a varied set of video-game maps.  This has served as the traditional test bed for subgoaling real-time search~\cite{bulitko08-dlrta-jair,bulitko10-knnLRTA-jair,hcdps-ajc}.  This examination demonstrates a statistical correlation between solution suboptimality and complexity measure values.  It also shows that machine learning can be used to predict  performance and facilitate algorithm parameterization.

We continue with an examination of algorithm performance beyond video-game pathfinding.  This study is performed using mazes and road maps.  To our knowledge, this is the first time that these algorithms have been applied to these domains.  These additional search spaces represent an incremental step towards more general spaces, and introduce several of the challenges that must be addressed if contemporary real-time algorithms are to be successfully adapted for broader domains such as planning.


\section{Problem Formulation}
\label{chap:formulation}

In this section we provide definitions for the terminology that will be used throughout the remainder of this paper.  We formally define heuristic search problems, and present our methodology for measuring the effectiveness of heuristic search algorithms.


\subsection{Heuristic Search Problems}

We define a \emph{search space} as follows: $S$ is a finite set of vertices or states; $E$ is a finite set of transitions, or weighted  edges, on $S \times S$. Edge weights are also known as edge costs and are all positive. Search space size is defined as the number of states $|S|$. In a search space, a {\em search problem} is defined as a pair of states: $s_\text{start} \in S$ is the start state and $s_\text{goal} \in S$ is the goal state.  We only consider search problems with a single goal state, since most video-game pathfinding is performed with a single goal state.

The agent's goal is to find a path (i.e., a sequence of edges) between the start and the goal state. Path cost is the sum of its edge costs and the agent attempts to find a path with the lowest possible cost (i.e., an optimal path between the start and the goal states). 

The search space comes with a heuristic function $h$ which any agent can use to guide its search. For a pair of arguments $s_1,s_2 \in S$ the heuristic function estimates the cost of a shortest path between $s_1$ and $s_2$. An optimal heuristic function $h^*$ gives the cost of an optimal path between two states. A heuristic function $h$ is \emph{admissible} if  $\forall s \in S \left[ h(s,s_\text{goal}) \leq h^*(s,s_\text{goal}) \right]$.

A heuristic search algorithm that is guaranteed to find a solution if one exists is called \emph{complete}.  An algorithm that is guaranteed to find an optimal solution is called an \emph{optimal} algorithm.  \emph{Learning} algorithms are those which make updates to their heuristic function during execution. 


\subsection{Search Performance}

We will use two major metrics to assess the performance of a heuristic search algorithm in this paper: solution suboptimality and pre-computation time.  These measures are collectively referred to as \emph{search performance}.  We define \emph{solution suboptimality} as the ratio of the cost of a solution to the cost of the optimal solution.  For example, if a given solution has cost $5$ and the optimal solution has cost $4$, then the solution suboptimality is $1.25$.

Most of the algorithms in this paper construct a database for the given search space.  Thus we measure \emph{database construction time} or \emph{pre-computation time}.  Any such preparatory behaviour performed by a search agent is referred to as being \emph{offline}, whereas any work done during active solving of specific search problems is referred to as being \emph{online}.  The comparative online and offline time consumption of the algorithms presented here has been previously studied by Bulitko et al ~\cite{bulitko10-knnLRTA-jair} and Lawrence et al.~\cite{hcdps-ajc}

\section{Real-time Search Algorithms}
\label{chap:algs}

Real-time search is a subclass of agent-centered search~\cite{Koenig:00c}.  In real-time search, the agent occupies a single state on the map at all time. In its current state, it performs planning to select a neighboring state. It then executes the move and makes the neighbor its new current state.  The amount of planning time per move is upper-bounded by a constant which is independent of the number of states. Every search decision by a real-time agent must be made within this fixed amount of time.  

Planning is typically done by expanding a \emph{local search space (LSS)}, a collection of nearby states, for consideration.  In an execution phase, the agent moves along the selected edge to the neighbour state.  The current state is updated to be the neighbour state, and a new planning step begins. Because an entire solution usually cannot be constructed within a single planning step, real-time heuristic search algorithms frequently make suboptimal moves, thereby increasing solution suboptimality. Thus a major line of research is improving the quality of planning and, subsequently, the resulting moves.

Recent real-time search algorithms attempt to do so by augmenting their online planning with the offline pre-computation of a search-space-specific database.  These databases are  used to provide one or several intermediate goals, or \emph{subgoals}, for use during search.  When an appropriate subgoal is found, search is directed towards that state rather than the original  goal.  This approach tends to improve search performance as a heuristic function is typically more accurate for a near-by subgoal than it is for a distant original goal.

We refer to the class of algorithms that are not real-time as \emph{conventional} heuristic search methods.  Popular search methods in this class include both optimal algorithms such as A*~\cite{Hart_Nilsson_Raphael_1968} and IDA*~\cite{Korf85depth-firstiterative-deepening}, and suboptimal algorithms such as HPA*~\cite{hpa:3225118} or PRA*~\cite{Sturtevant:05b}.  We do not focus on conventional heuristic search, but discuss some of them briefly.


\subsection{LRTA*: The Foundation}
\label{sec:lrta}

Learning real-time A* \emph{(LRTA*)}~\cite{Korf-AIJ} is the first real-time heuristic search algorithm we discuss.  LRTA* serves as the foundation for three of the subsequent real-time search algorithms discussed in this section.

\begin{algorithm}[htbp]
\caption{LRTA*($s_\text{start},s_\text{goal},d$)}
\label{alg:lrta}
\begin{algorithmic}[1]
\STATE $s \leftarrow s_\text{start}$ \label{alg:lrta:init}
\WHILE { $s \neq s_\text{goal}$ }  
\STATE expand LSS of $s$ \label{alg:lrta:expand}
\STATE find a frontier  $s'$ minimizing $g(s,s') + h(s',s_\text{goal})$ \label{alg:lrta:eval}
\STATE $h(s,s_\text{goal}) \leftarrow g(s,s') + h(s',s_\text{goal})$ \label{alg:lrta:learn}
\STATE change $s$ one step towards $s'$ \label{alg:lrta:move}
\ENDWHILE
\end{algorithmic}
\end{algorithm}

Pseudo-code for LRTA* is given as Algorithm~\ref{alg:lrta}.  The agent begins by initializing its current location $s$ to the start state $s_\text{start}$ (line~\ref{alg:lrta:init}).  Next, a set of successor states up to $d$ moves away is generated surrounding $s$ (line~\ref{alg:lrta:expand}).  Once the local search space (LSS) is expanded, its frontier (i.e., border) states are considered and the most promising state $s'$ is selected in line~\ref{alg:lrta:eval}.  The chosen state is that which minimizes the function $g(s,s') + h(s',s_\text{goal})$, where $g(s,s')$ is the cost of an optimal path from $s$ to $s'$ within the LSS.  To deter state revisitation, the heuristic value $h(s,s_\text{goal})$ is updated to $g(s,s') + h(s',s_\text{goal})$ (line~\ref{alg:lrta:learn}).  The agent then moves one step along the path towards $s'$ (line~\ref{alg:lrta:move}). These steps are repeated until the agent's current state matches the goal state.  LRTA* is only complete given an admissible heuristic.

Despite the learning step in line~\ref{alg:lrta:learn}, LRTA* is prone to frequent state revisitation.  This tendency has been named \emph{scrubbing}~\cite{knnLRTA-aiide09}, since the agent appears to ``scrub'' back and forth over small regions of the search space to fill in heuristic depressions~\cite{Ishida:1997:RSL:275618}.  Scrubbing behaviour is detrimental for two  reasons.  First, state revisitation necessarily increases suboptimality of solutions.  Second, scrubbing in applications such as video-game pathfinding is visually unappealing and reduces player immersion.  This is a major barrier preventing LRTA* from being applied in commercial video games.


\subsection{Hill-climbing}
\label{sec:hc}

We define a \emph{hill-climbing} (HC) agent as a greedy LRTA*-like agent which performs no heuristic updates and only uses the immediate neighbours of the current state in the LSS.  The agent will move to the neighbour state with the lowest heuristic value.  Hill-climbing is not complete.  Search is terminated if the agent ever reaches a state with a heuristic value less than or equal to all surrounding states or after a certain quota on the number of steps is reached.  To detect search problems where one state is HC-reachable from another, that is, where a hill-climbing agent will find a solution, we use Algorithm~\ref{alg:hcReachable}~\cite{bulitko10-knnLRTA-jair}.  The algorithm returns true if the state $s_2$ can be hill-climbed to from the state $s_1$ in no more than $b$ moves. Here and below when we talk about hill-climbability, we always enforce a limit on the number of moves.

\begin{algorithm}[htbp]
\caption{HC-Reachable($s_1,s_2,b$)}
\label{alg:hcReachable}
\begin{algorithmic}[1]
\STATE $s \leftarrow s_1$
\STATE $i \leftarrow 0$
\WHILE { $s \neq s_2  \And i < b$ }  
\STATE generate immediate neighbours of $s$
\IF {$h(s,s_2) \leq h(s',s_2)$ for any neighbour $s'$}
\RETURN false
\ENDIF
\STATE set $s$ to the neighbour with the lowest $g(s,s') + h(s',s_2)$
\STATE $i \leftarrow i+1$
\ENDWHILE
\RETURN $s = s_2$
\end{algorithmic}
\end{algorithm}
\vspace{-0.4cm}


\subsection{D LRTA*}

Dynamic LRTA* (\emph{D LRTA*})~\cite{bulitko08-dlrta-jair} was designed to mitigate the scrubbing behaviour of LRTA*.  Two improvements are made over the original LRTA*: dynamic selection of search depth $d$, and case-based subgoaling.  Both of these ends are accomplished with the aid of a pre-computed database generated offline, before the D LRTA* agent is tasked with solving search problems online.

The subgoal database is computed as follows. First, the search space is partitioned into regions by repeatedly merging cliques of states~\cite{Sturtevant:05b}. Then a representative state is randomly picked within each partition. For each pair of the partitions $A$ and $B$, an optimal path is computed between their representatives $a$ and $b$ using A*. The first state on the path that leaves the start partition $A$ is stored in the database as a subgoal for the pair of partitions: $s_{AB}$.

Online, D LRTA* uses LRTA*, except instead of computing the heuristic with respect to the global goal $s_\text{goal}$ it computes it with respect to a subgoal. Specifically, suppose the agent's current state is in a partition $A$ and the global goal is in the partition $B$. Then the agent will compute its heuristic with respect to the subgoal $s_{AB}$ and head towards it. As soon as it leaves $A$ and enters $C$, it will change its subgoal to $s_{CB}$ and so on.


\subsection{kNN LRTA*}

While D LRTA* uses LRTA* online, the scrubbing behavior is reduced and solution quality is increased due to the use of nearby subgoals in place of a distance global goal. However, repeated clique abstraction results in complex partitions and, as a result, the next subgoal may not be reachable without scrubbing {\em within} a partition. Additionally, the entire space needs to be partitioned and the partitions stored in memory. 

kNN LRTA*~\cite{bulitko10-knnLRTA-jair} aims to address both issues. First it makes sure that the next subgoal, if it exists in its database, is reachable from the current state without scrubbing. Second, it avoids a full enumeration of all states, inherent to partitioning. Instead of partitioning the entire search space and then computing a subgoal for each pair of partition representatives, kNN LRTA* computes subgoals for random problems. Specifically, offline it optimally solves $N$ random problems in the search space using A*. Each resulting optimal path is compressed into a series of subgoals such that each subgoal is hill-climable (Algorithm~\ref{alg:hcReachable}) from the previous one. Each such sequence of subgoals is stored in the database together with the corresponding start and goal states.

Then online, a kNN LRTA* agent searches through its database to find a problem most similar to the problem at hand. The similarity is defined as the sum of the heuristic distances between the start states and between the goal states of the search problem and a candidate database entry.  The agent makes sure that the start state of the database problem is HC-reachable from its start state. It also makes sure that it will be able to hill-climb from the goal state of the database problem to its goal state. If such a suitable database record exists that the agent hill-climbs from its start state to the problem's start state, then uses LRTA* to traverse the series of subgoals, eventually reaching the database problem's goal state. Then it hill-climbs to the actual goal state. 

If no hill-climable problem exists in the database then the agent falls back onto LRTA* with respect to the global goal.


\subsection{HCDPS}
\label{sec:hcdps}

kNN LRTA* performs no scrubbing when the problem at hand has a suitable record in the database. However, as the database is comprised of solutions to random problems, there is no guarantee that an on-line problem will have a suitable record. When such a record is missing, kNN LRTA* degenerates to LRTA* which results in possibly extensive scrubbing and poor solution quality.

Hill-Climbing Dynamic Programming Search (HCDPS)~\cite{hcdps-ajc} computes its subgoal database more systematically, thereby ensuring that each online problem has a suitable record. Unlike D LRTA*, however, it eschews the clique abstraction in favor of hill-climbing-friendly partitioning.

The first step in this database construction involves partitioning the entire search space into a set of \emph{HC regions}.  Each HC region is a set of states, with one state designated as the \emph{seed state}.  Every state in a HC region is mutually HC-reachable with the seed state. The regions are constructed by randomly selecting states from the yet unpartitioned set of states and growing regions from each state in a breadth-first fashion. A state is added to the region if it neighbours a state already in the region and is mutually HC-reachable with the seed state.

The second step in the database construction involves computing a path between the representative seed states of each pair of HC regions.  This path is then converted into a chain of subgoals and stored as a database record in the same way as with kNN LRTA*.

To speed up database pre-computation, HCDPS does not calculate the true optimal path between every pair of representative seed states.  Instead, it uses dynamic programming to assemble such paths by chaining together optimal paths between seed states of neighbouring HC regions via dynamic programming.

When presented with a problem to solve online, HCDPS looks up the HC regions for the start and the goal states of the problem and retrieves a chain of subgoals connecting the seed states of the the HC regions. By construction of HC regions, it is able to hill-climb from the start state at hand to the seed state of the first HC region. Then it hill-climbs between subgoals in the record, reaching the seed state in the final HC region. From there it hill-climbs to the global goal state. Due to the guaranteed hill-climability, HCDPS forgoes LRTA* and does not perform any heuristic learning  at all.


\subsection{TBA*}
\label{sec:tba}

The final real-time algorithm we discuss in this paper is \emph{time-bounded A*} (TBA*)~\cite{DBLP:conf/ijcai/BjornssonBS09}.  Unlike the previous three algorithms discussed, TBA* does not make use of a subgoal database. We are including it in our analysis to show the generality of our techniques.

TBA* is essentially a time-sliced A*. It builds its open and closed lists from the start state in the same way as A*. Unlike A*, it does not wait for the goal state to be at the head of its open list. Instead, after a fixed number of expansions, a TBA*-driven agent takes one step towards the most promising state on the open list. If, while on its way to the most promising open-list state, such a state changes, the agent re-plans its path towards the new target. All of the planning operations are time-sliced so that the resulting algorithm is real-time.

\section{Related Work} 
\label{chap:related}

There has been significant past research on analyzing search space complexity and prediction of search algorithm performance.  A small subset of this work has focused on real-time heuristic search in particular.  In this section we explore some of this related work.

\subsection{Real-time Heuristic Search Analysis}

Koenig first presented motivation for analysis of real-time search performance~\cite{koenig98}.  He indicated that, unlike conventional search methods, real-time search was poorly understood.  As a preliminary effort, he discussed the impact of domain, heuristic and algorithm properties on search behaviour, noting that real-time search and conventional search are affected quite differently by these factors.  Specifically, Koenig stated the following: \begin{quote} In general, [...] not much is known about how domain properties affect the plan-execution time of real-time search methods, and there are no good techniques yet for predicting how well they will perform in a given domain.  This is a promising area for future research. \end{quote}

Similarly, Koenig stated that there were no strong methods for predicting the comparative performance of multiple different real-time search algorithms on a given planning task.  As an example, he compared the disparate performances of LRTA* and the similar algorithm \emph{Node Counting} on a selected set of domains.  Koenig observed that, despite comparable typical case performance over thousands of trials, worst case solution costs for Node Counting are substantially more expensive than for LRTA*.  Furthermore, he indicated the difficulty of predicting these degenerate cases of Node Counting.

Koenig's analysis of real-time search was limited to LRTA* and a few variants.  We seek to extend analysis to the more contemporary class of database-driven algorithms discussed in Section~\ref{chap:algs}.  The major motivation that we take from Koenig's work is that real-time search behaviour differs greatly not only from conventional search, but also among different real-time algorithms and search spaces.  Therefore, our system for characterizing algorithm performance discussed in Section~\ref{chap:complex} is designed specifically with properties of contemporary real-time algorithms in mind.

Bulitko and Lee performed a large scale analysis of numerous real-time heuristic search algorithms, including LRTA*, $\epsilon$-LRTA*, SLA* and $\gamma$-trap~\cite{DBLP:journals/jair/BulitkoL06}.  They developed LRTS, a unified framework for these four algorithms, and performed a large scale empirical study across several search spaces.  As motivation, they cited the present difficulty of appropriately selecting algorithms and parameters from the available pool. Four of the five real-time search algorithms that we explore in detail have been developed in the time after the work of Bulitko and Lee.  We therefore find ourselves again faced with a similar problem of an abundance of algorithms of which the relative merits have only been briefly explored in a small selection of search spaces~\cite{hcdps-ajc}.  While we do not take the approach of combining the algorithms into a single framework, we do share the motivation of facilitating algorithm selection and parameterization.


\subsection{Characterizing Search Spaces}

Our decision to analyze real-time search performance via search space features is largely motivated by the disparate performance in initial experiments applying recent real-time methods to a wider variety of search spaces.  Understanding search spaces can serve two important purposes.  First, we can more make more informed decisions when selecting or designing algorithms for a given search space.  Second, we can more consistently compare the performance of new algorithms by establishing benchmark search spaces with well understood characteristics.

Two of the features that we present in Section~\ref{chap:complex} are derived from the work of Ishida.  Ishida identified that given the necessarily committal behaviour of real-time search algorithms, they are more susceptible to local heuristic topography than conventional search methods~\cite{Ishida:1997:RSL:275618}.  To measure this topography empirically, Ishida provides the following definitions: \begin{quote} A \emph{heuristic depression} is a set of connected states with heuristic values less than or equal to those of the set of immediate and completely surrounding states.  A heuristic depression is \emph{locally maximal}, when no single surrounding state can be added to the set; if added, the set does not satisfy the condition of a heuristic depression.~\cite{Ishida:1997:RSL:275618} \end{quote}  As pointed out in Section~\ref{chap:algs}, heuristic depressions can cause scrubbing to occur in LRTA*-based search methods.  Ishida performed experiments that enumerated heuristic depressions in mazes and sliding tile puzzles.  He also provided intuition as to how these search space features might affect real-time search performance in terms of the number of heuristic updates performed.

Ishida also conducted an empirical comparison of the performance of LRTA* and two variants, \emph{Real-Time A*} (RTA*)~\cite{Korf-AIJ} and \emph{Local Consistency Maintenance} (LCM)~\cite{Pemberton92makinglocally}.  The analysis focused on comparing the learning efficiency of these algorithms when multiple trials are allowed (i.e., as the heuristic converges) or when different initial heuristic functions are used.

More recently, Hoffman extended a similar study of heuristic topology\footnote{The terms topography and topology are used interchangeably in the literature when discussing heuristic functions.} to general planning~\cite{DBLP:conf/ijcai/Hoffmann01,DBLP:conf/aips/Hoffmann02}.  On a set of $13$ well-established planning benchmarks, he enumerated topological heuristic features and domain properties to sort the benchmarks into a complexity taxonomy.  This is similar to our ranking of search spaces based on complexity in Sections~\ref{chap:videogames} and~\ref{chap:combin}, although Hoffman does not compute empirical correlations to performance.  Hoffman concluded that the benchmark taxonomy would shed insight on the levels of success of heuristic search planning in these benchmarks.  He also claimed that it could inform subsequent improvement of heuristic functions, allow prediction of planning performance and assist in developing more  challenging benchmark sets.

Another of the complexity measures we use to characterize search spaces is taken from the research of Mizusawa and Kurihara~\cite{DBLP:journals/heuristics/MizusawaK10}.  They successfully demonstrated a  link between search performance in gridworld pathfinding domains and two ``hardness measures'': initial heuristic error and probability of solution existence.  They define initial heuristic error for a search problem as $E = \sum_{s\in S'} h^*(s) - h_0(s)$ where $S'$ is the set of all states on some path connecting the start and goal states.

The search spaces used by Mizusawa and Kurihara are generated randomly by making random cells in the gridworld untraversable.  The percentage of untraversable cells is called the \emph{obstacle ratio}, $p$.  Since the obstacles are placed randomly, solutions are not guaranteed to exist for search problems.  The entropy $H = -p \log _2 p - (1-p) \log _2 (1-p)$ is used as their measure of likeliness of solution existence.
Mizusawa and Kurihara demonstrated that $E$, $H$, and LRTA* and RTA* solution cost are all maximized at a similar obstacle ratio of approximately $41\%$ in square, randomly generated maps.  Unlike Mizusawa and Kurihara, the search problems we consider are guaranteed to have solutions.  However, their system of using complexity measures to characterize search spaces with respect to search performance was informative to our own research.


\subsection{Predicting Search Performance}

Previous work has been conducted to predict the performance of simpler real-time heuristic search algorithms.  Citing important applications in planning (e.g., as part of the Heuristic Search Planner~\cite{Bonet01planningas} and the Fast-Forward Planner~\cite{Hoffmann01ff:the}), L\`{o}pez sought to model the efficiency of heuristic hill-climbing by modelling the algorithm as a Markov process~\cite{DBLP:conf/aimsa/Lopez08}.  Using several sizes of the sliding tile puzzle, the model could reasonably predict the likelihood that a hill-climbing agent reaches a target state in a number of moves equal to the initial heuristic value.  L\`{o}pez stated that this model is useful not only as a predictor of simple real-time search performance, but as a gauge of heuristic accuracy.

One of the complexity measures we present in Section~\ref{chap:complex} bears similarity to L\`{o}pez's work.  We consider the probability with which a hill-climbing agent successfuly reaches a target state.  Unlike L\`{o}pez, we do not differentiate between cases where the path taken by the agent is more or less expensive than the heuristic estimate.

Recent research by Kotthoff et al.~\cite{KotthoffGM11} performed a preliminary assessment of the suitability of machine learning as a predictor of search algorithm performance.  They discussed several machine learning techniques for predicting which search algorithm from a collection, or portfolio, would have the best performance for a given search problem.  Their method uses past algorithm performance and search space features as training data for their models, similar to our use of complexity measures as input to a machine learning predictor presented in Section~\ref{chap:predictive}.  However, to our knowledge, we present the first exploration of machine learning to predict the performance of a collection of real-time heuristic search algorithms.


\section{Complexity Measures} 
\label{chap:complex}

To prepare for the application of recent real-time heuristic search to new domains, we sought to first find a method of empirically characterizing the challenges that new search spaces would bring.  Intuitively, a maze is more ``complex'' than an open room for an agent to navigate, but this notion of complexity is not as evident in less visual search spaces.

Our goals for this research are two-fold.  We first seek to facilitate algorithm selection.  Second, we wish to build an understanding of search space features that will inform subsequent algorithm development.


\subsection{Domain-Independent Complexity Measures}

We use a set of eight domain-independent complexity measures.  All of the measures are calculated independently of any algorithm, and may thus be useful for assessing the suitability of several different algorithms for a particular search space.  For presentational clarity, we discuss the measures as they are calculated for search spaces with a single goal state.  However, all of the measures could be easily adapted to serve search spaces with multiple goals.

\begin{enumerate}

\item{\textbf{HC Region Size} $\in [1,\infty)$ is the mean number of states per abstract region when the search space is partitioned using the abstraction method of HCDPS   (Section~\ref{sec:hcdps}).}

\item{\textbf{HC Probability} $\in [0,1]$ is the probability that a randomly selected state is HC-reachable from another randomly selected state.  HC-reachability is checked using Algorithm~\ref{alg:hcReachable} from Section~\ref{sec:hc}.}

\item{\textbf{Scrubbing Complexity} $\in [1,\infty)$ is the mean number of visits among all states receiving at least one visit in the solution returned by an LRTA* agent.  This measure is intended to model real-time search behaviour when no sub-goals are made available to the agent.}

\item{\textbf{Path Compressibility} $\in [1,\infty)$ is the mean number of subgoal states when the solution to a random search problem is compressed using the compression schema of kNN LRTA*/HCDPS.}

\item{\textbf{A*-Difficulty} $\in [1,\infty)$ is the mean number of states on the A* closed list after solving a random search problem, scaled by the length of the solution.  This measure has been previously used to measure the complexity of search problems for conventional search.}

\item{\textbf{Heuristic Error} $\in [0,\infty)$ is the average cumulative difference in value between $h_0$ and $h^*$ across all reachable states sampled over a random set of goal states~\cite{DBLP:journals/heuristics/MizusawaK10}.}

\item{\textbf{Total Depression Width} $\in [0,\infty)$ is the mean number of depressed states for a random goal state.  States within the heuristic depression containing the goal state are excluded from consideration, since that depression is not inhibitive for real-time search.  This measure is intended to model the likelihood that a real-time agent will become temporarily trapped during search.}

\item{\textbf{Depression Capacity} $\in [0,\infty)$ is the mean sum of depths across all depressed states for a random goal state.  Again, the heuristic depression containing the goal state is not considered.  This measure is intended to model the duration for which the agent will be trapped in a depression.}

\end{enumerate}

\subsection{Computing Complexity Measures in Practice}

There are two  considerations when computing the values of these measures for a search space.  First, we must make sure that we are sampling across a sufficient number of goal states or search problems for the measured value to be representative of the search space.  Second, we must calculate the measures in a way that avoids introducing bias towards search space size.  In this section we discuss how we address both concerns.

\bigskip\subsubsection{Sufficient Sampling}

To determine an appropriate sample size to calculate each measure, we perform repeated sampling until the current observed mean becomes relatively stable.  As an aid we used \emph{stability graphs}, which plot the current sample number against the current observed mean.  After a certain number of samples, we observe a decrease in fluctuation of the current mean value.  This number of samples is then used to compute that complexity measure in practice.

There is no constant sample size that will be sufficient for all search spaces. Larger search spaces, and those with diverse regional complexity, will require a larger number of samples.  Thus, this method serves as a reasonable guide for selecting appropriate sample sizes.

\bigskip\subsubsection{Localized Complexity}
\label{sec:LocComp}

The ultimate purpose of the complexity measures is to allow a meaningful comparison of search spaces with different structure. Thus, to ensure that the measures are effectively capturing local complexity rather than being confounded by search space size, we  constrain the calculation of the complexity measures to a fixed portion of the search space.  To accurately reflect the entirety of the search space, we  then  repeatedly sample across several such random portions of the search space.  The random sub-spaces are generated via a bounded breadth-first search originating at a randomly selected state.  After repeating the sampling across a variety of bounded regions, we compute the aggregate complexity measure as the mean of the collected measure values for the sampled regions.  The subspaces are of the same size for all of the search spaces being compared in this way.

Aside from preventing a bias to search space size, this sampling technique can improve the efficiency of calculating the complexity measures.  This is of particular importance when applying the measures to large search spaces.  For example, calculating scrubbing complexity on a large video-game map can be very expensive, since solving even a single instance of LRTA* takes a non-trivial amount of time.  In this approach we instead solve a larger number of LRTA* problems in multiple smaller sub-spaces.

The final benefit of this technique is that it allows the complexity measures to be computed for implicitly defined search spaces which are too large to fit in memory in their entirety. Specifically, instead of expanding the entire search space at once, we expand only one bounded portion at a time.  The complexity measures are computed for that bounded portion, and the process is repeated.

\section{Rank Correlation}
\label{chap:videogames}


In this section we present experimental evidence linking the performance of real-time heuristic search algorithms on video-game pathfinding to the values of the complexity measures presented in Section~\ref{chap:complex}.  We begin by describing our selection of search spaces for experimentation.  We then detail our methodology for measuring search performance.

We make use of two distinct methods for linking the complexity measures to search performance.  The first method is to compute the rank correlation between the values of a complexity measure and the performance of an algorithm for a given search space.  This method allows us to identify which search space features can significantly impact the performance of an algorithm.  This is useful for understanding the limitations of current algorithms, and for gauging the relative difficulty of search spaces for an algorithm.

Our second method involves using machine learning to build predictors of search space performance.  This demonstrates that we can use the values of the complexity measures to assist in algorithm selection and parameterization.  The second method is explored in Section~\ref{chap:predictive}.


\subsection{Pathfinding as a Search Problem}

Video-game pathfinding has been been the traditional testbed for the recent real-time heuristic search algorithms we examine.  Pathfinding in commercial video games is very resource limited, often limited to a specific amount of time (often less than $1$ ms)~\cite{bulitko10-knnLRTA-jair}.  It is therefore a natural application for real-time search.  In this section we formally describe video-game pathfinding as a search problem as defined in Section~\ref{chap:formulation}.

A video-game pathfinding search space, or \emph{map}, consists of a grid of cells which are either open or occupied by an obstacle. Each open cell is a state in the search space. Immediately adjacent cells are connected by edges. Each cell has a maximum of eight neighbors: four in the cardinal directions (north, south, east, west) and four in the diagonal directions. The cost of every cardinal edge is $1$, and the cost of every diagonal edge is $1.4$.\footnote{We follow the common practice of using $1.4$ instead of $\sqrt{2}$ for diagonal costs. This is done to avoid rounding errors.} We use \emph{octile distance} as our initial heuristic function.  The octile distance between two states is defined as $| \Delta x - \Delta y | + 1.4 \min\{\Delta x, \Delta y \}$ where $\Delta x$ and $\Delta y$ are the absolute values of the differences in $x$ and $y$ coordinates of the two states.


\subsection{Experimental Design}
\label{sec:mapExpDesign}

\begin{figure*}[htbp]
\centering
\includegraphics[height=4.1cm]{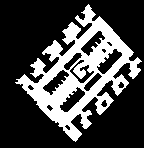}
\hspace{0.3cm}
\includegraphics[height=4.1cm]{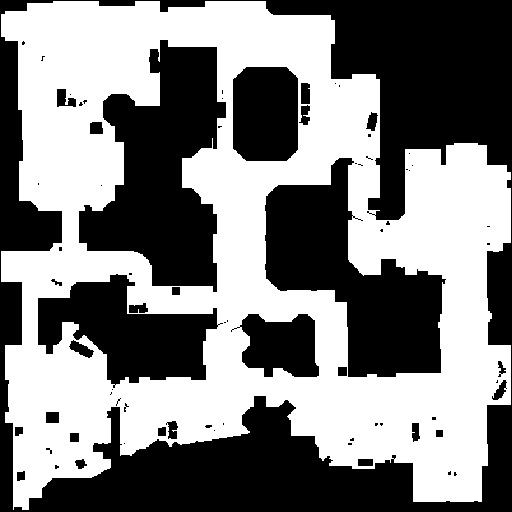}
\hspace{0.3cm}
\includegraphics[height=4.1cm]{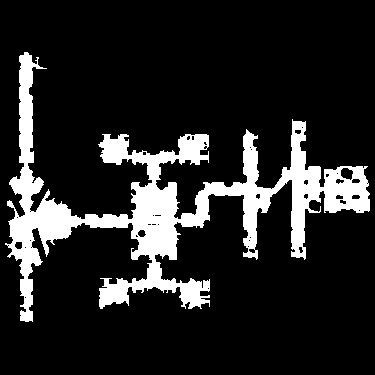}
\hspace{0.3cm}
\includegraphics[height=4.1cm]{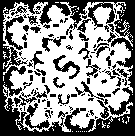}
\caption{Sample maps from the video games {\em Baldur's Gate}, {\em Counter-strike: Source}, {\em Dragon Age: Origins} and {\em Warcraft 3}.} \label{fig:maps}
\end{figure*}

To establish a quantifiable link between the complexity measures presented in Section~\ref{chap:complex} and the performance of real-time heuristic search, we first conducted experiments in pathfinding on $20$ video-game maps, with five from each of {\em Baldur's Gate}~\cite{BaldurGate-short}, {\em Counter-strike: Source}~\cite{CSS-short}, {\em Warcraft 3}~\cite{Warcraft3-short} and {\em Dragon Age: Origins}~\cite{DragonAge-short}.  Maps from the first three games have been previously used in recent real-time heuristic search literature~\cite{bulitko10-knnLRTA-jair, hcdps-ajc}.  Most of these maps are available online through Nathan Sturtevant's Moving AI Lab~\cite{movingAI}.  Four example maps are presented in Figure~\ref{fig:maps}.

As discussed in Section~\ref{chap:formulation}, we consider two basic measures of a real-time search algorithm's performance: solution suboptimality and database computation time.\footnote{Since LRTA* and TBA* do not utilize subgoal databases they are omitted from our discussion of precomputation time.}

In using our localized complexity technique (Section~\ref{sec:LocComp}), we first scaled all of the maps to approximately $50$ thousand states.  We then generated $10$ sub-maps per map by randomly selecting samples of exactly $20$ thousand states from each map.  Sub-maps were generated via breadth-first search, originating at a random state in the search space.  We treated each of these $20 \times 10 = 200$ sub-maps as a whole and distinct search space for our experiments by removing all states not reached via the breadth-first search.  An additional benefit of this approach was to increase the number of available data points $10$-fold, while retaining maps that looked similar to full-size game maps, and yet diverse enough to establish trends in search space complexity.

To compute the solution suboptimality of each algorithm, we generated a random set of $250$ pathfinding problems on each sub-map, and solved each problem with each of the three algorithms.  This number of problems was chosen to cover a sufficient sample of typical search problems while remaining tractable.  The problems were constrained to have solutions of length at least $10$ to exclude trivial problems.

\subsection{Algorithm Implementation Details}

In this section we examine five real-time heuristic search algorithms: LRTA*, D LRTA*, kNN LRTA*, HCDPA and TBA*.  All of our algorithm implementations exist within a common framework.  This helps reduce any discrepancies in performance due to differences in tie-breaking procedures, data structures used, etc.

In order to make the trends in algorithm performance most visible, we have selected a combination of algorithm parameters and search problems such that each algorithm is faced with both easy and difficult search problems.  Our objective was not to hand-tune parameters for the best performance, but to yield a wide spread of performance.  LRTA* was run with a lookahead depth of $d = 1$.  D LRTA* ran with $\ell = 5$. kNN LRTA* used $N = 1000$ and $M = 10$. HCDPS ran with $r=1$.  All on-line HC checks were limited to $250$ steps.  TBA* was run with a resource limit of $R=5$.


\subsection{Rank Among Measures and Algorithm Performance}

To empirically link the complexity measures to search performance, we calculated the Spearman rank correlation coefficients between the average solution suboptimality and a complexity measure's values~\cite{spearman1904}.  The coefficient reflects the tendency for two variables to increase or decrease monotonically together, possibly in a non-linear fashion.  The ability to gauge non-linear correlation was our main motivation for selecting Spearman correlation over other measures such as Pearson correlation.

Let $X$ and $Y$ be two sets of data.  Let $x_i$ be the rank, or position in descending order, of the $i^{\text{th}}$ element in $X$.  Define $y_i$ analogously.  Then Spearman correlation is defined as $$ \text{corr}(X,Y) = \dfrac{\sum_i(x_i - \bar{x})(y_i - \bar{y})}{\sqrt{\sum_i(x_i - \bar{x})^2\sum_i(y_i - \bar{y})^2}} $$ where $\bar{x}$ is the mean rank for all $x_i$, and $\bar{y}$ is the mean rank for all $y_i$.

We computed two sets of correlations: $\rho_\text{mean}$ using mean suboptimality, and $\rho_\text{median}$ using median suboptimality.  We used median suboptimality to mitigate the impact of outliers: rare but vastly suboptimal solutions.  This allowed us to understand the differences between typical and outlying cases by contrasting the mean and median cases.

For ease of reading, we have color-coded the reported correlations in four bins: $\color{Brown}[0, 0.25)$, $\color{RedOrange}[0.25,0.50)$, $\color{YellowOrange}[0.50,0.75)$ and $\color{OliveGreen}[0.75,1.00]$.  These bins have no empirical significance, so the coloration of the tables should not stand as a replacement for a full reading of the data.

\subsection{Correlation Among Algorithms}

Our first step was to test our hypothesis that some real-time search algorithms respond differently to certain search space features than others.  In other words, we wanted to demonstrate that search problems can not simply be sorted on a one-dimensional continuum of complexity.  A certain search problem may be very difficult for one algorithm, and yet comparatively easy for another.

To demonstrate this, we calculated the Spearman rank correlation between the solution suboptimality for each of the five algorithms in a pairwise fashion.  The correlations between the mean suboptimalities are presented in Table~\ref{tab:corrAlgsMean} and between the median suboptimalities in Table~\ref{tab:corrAlgsMedian}.  The marker $\dagger$ indicates statistical {\em in}significance (i.e., $p > 0.05$). We chose to use a marker for statistical insignificance simply because most of our results are significant at the level of $p \le 0.05$ and marking them all would clutter the tables.

\begin{table}[htbp]\small
\center
\begin{tabular}{ l | c | c | c }
& LRTA* & D LRTA* & kNN LRTA* \\ \hline \hline
LRTA* & -- & $\color{RedOrange}0.4862$ & $\color{Brown}0.182$ \\
D LRTA* & $\color{RedOrange}0.4862$ & -- & $\color{RedOrange}0.263$ \\
kNN LRTA* & $\color{Brown}0.182$ & $\color{RedOrange}0.263$ & -- \\
HCDPS & $\color{Brown}0.140$ & $\color{Brown}0.163$ & $\color{YellowOrange}0.538$ \\
TBA* & $\color{OliveGreen}0.857$ & $\color{RedOrange}0.425$ & $\color{RedOrange}0.284$ \\
\end{tabular}

\vspace{0.4cm}

\begin{tabular}{ l | c | c }
& HCDPS & TBA* \\ \hline \hline
LRTA* & $\color{Brown}0.140$ & $\color{OliveGreen}0.857$ \\
D LRTA* & $\color{Brown}0.163$ & $\color{RedOrange}0.425$ \\
kNN LRTA* & $\color{YellowOrange}0.538$ & $\color{RedOrange}0.284$ \\
HCDPS & -- & $\color{Brown}0.210$ \\
TBA*  & $\color{Brown}0.210$ & -- \\
\end{tabular}
\caption{Correlation among algorithms' mean suboptimality.} \label{tab:corrAlgsMean}
\vspace{-0.5cm}

\end{table}

\begin{table}[htbp]\small
\center
\begin{tabular}{ l | c | c | c }
& LRTA* & D LRTA* & kNN LRTA* \\ \hline \hline
LRTA* & -- & $\color{Brown}-0.154$ & $\color{YellowOrange}0.714$ \\
D LRTA* & $\color{Brown}-0.154$ & -- & $\color{Brown}0.071 \dagger$ \\
kNN LRTA* & $\color{YellowOrange}0.714$ & $\color{Brown}0.071 \dagger$ & -- \\
HCDPS & $\color{OliveGreen}0.763$ & $\color{Brown}-0.018 \dagger$ & $\color{OliveGreen}0.821$ \\
TBA* & $\color{OliveGreen}0.764$ & $\color{RedOrange}-0.324$ & $\color{RedOrange}0.446$ \\
\end{tabular}

\vspace{0.4cm}

\begin{tabular}{ l | c | c }
& HCDPS & TBA* \\ \hline \hline
LRTA* & $\color{OliveGreen}0.763$ & $\color{OliveGreen}0.764$ \\
D LRTA* & $\color{Brown}-0.018 \dagger$ & $\color{RedOrange}-0.324$ \\
kNN LRTA* & $\color{OliveGreen}0.821$ & $\color{RedOrange}0.446$ \\
HCDPS & -- & $\color{YellowOrange}0.584$ \\
TBA*  & $\color{YellowOrange}0.584$ & -- \\
\end{tabular}
\caption{Correlation among algorithms' median suboptimality.} \label{tab:corrAlgsMedian}
\vspace{-0.5cm}
\end{table}

We observed very few strong correlations in the mean case.  The only strong correlation was between TBA* and LRTA* ($\rho = 0.857$), with a moderate correlation observed between kNN LRTA* and HCDPS ($\rho = 0.538$).

In contrast, we observed  stronger correlations in the median case.  HCDPS and kNN LRTA* ($\rho = 0.821$), HCDPS and LRTA* ($\rho = 0.714$), and kNN LRTA* and LRTA* ($\rho = 0.763$) are all substantially more correlated than in the mean case, and TBA* and LRTA* remain fairly correlated ($\rho = 0.764$).  By considering only median solution suboptimality, we effectively remove from consideration any degenerate cases where the algorithms return vastly suboptimal solutions.  Therefore, the algorithms are more frequently observed to achieve optimal or near optimal median solution costs, and the overall variance in median suboptimality is lower.  We hypothesize that this ``tightening of the pack'' is responsible for the higher correlations.  We  postulate that the relative disparities in algorithm performance manifest more apparently in outlying cases.


\subsection{Correlation Among Complexity Measures}
\label{sec:amongComp}

Note that some of the complexity measures are quite highly correlated.  Similar to the inter-algorithm comparison in the previous section, we present the pairwise Spearman rank correlation of all of the complexity measures in Table~\ref{tab:corrComp}.

\begin{table}[htbp]\scriptsize
\center
\begin{tabular}{ l | c | c | c }
& HC Region Size & HC Probability & Scrub. Complexity \\ \hline \hline
HC Region Size & -- & $\color{OliveGreen}0.759$ & $\color{Brown}-0.117 \dagger$ \\ 
HC Probability & $\color{OliveGreen}0.759$ & -- & $\color{YellowOrange}-0.529$ \\ 
Scrubbing Complexity & $\color{Brown}-0.117 \dagger$ & $\color{YellowOrange}-0.529$ & -- \\
Path Compressibility & $\color{YellowOrange}-0.697$ & $\color{OliveGreen}-0.920$ & $\color{YellowOrange}0.602$ \\
A* Difficulty & $\color{Brown}-0.055 \dagger$ & $\color{YellowOrange}-0.514$ & $\color{OliveGreen}0.918$ \\
Heuristic Error & $\color{Brown}-0.068 \dagger$ & $\color{YellowOrange}-0.540$ & $\color{OliveGreen}0.945$ \\
Depression Width & $\color{RedOrange}-0.435$ & $\color{YellowOrange}-0.611$ & $\color{YellowOrange}0.696$ \\
Depression Capacity & $\color{Brown}0.072 \dagger$ & $\color{Brown}-0.178$ & $\color{YellowOrange}0.633$ \\
\end{tabular}

\vspace{0.4cm}

\begin{tabular}{ l | c | c | c }
& Path Compressibility & A* Difficulty & Heuristic Error \\ \hline \hline
HC Region Size & $\color{YellowOrange}-0.697$ & $\color{Brown}-0.055 \dagger$ & $\color{Brown}-0.068 \dagger$ \\ 
HC Probability & $\color{OliveGreen}-0.920$ & $\color{YellowOrange}-0.514$ & $\color{YellowOrange}-0.540$ \\ 
Scrubbing Complexity & $\color{YellowOrange}0.602$ & $\color{OliveGreen}0.918$ & $\color{OliveGreen}0.945$ \\ 
Path Compressibility & -- & $\color{YellowOrange}0.613$ & $\color{YellowOrange}0.624$ \\ 
A* Difficulty &  $\color{YellowOrange}0.613$ & -- & $\color{OliveGreen}0.983$ \\ 
Heuristic Error & $\color{YellowOrange}0.624$ & $\color{OliveGreen}0.983$ & -- \\
Depression Width & $\color{YellowOrange}0.622$ & $\color{YellowOrange}0.621$ & $\color{YellowOrange}0.629$ \\
Depression Capacity & $\color{Brown}0.230$ & $\color{YellowOrange}0.594$ & $\color{YellowOrange}0.606$ \\
\end{tabular}

\vspace{0.4cm}

\begin{tabular}{ l | c | c }
& Depression Width & Depression Capacity \\ \hline \hline
HC Region Size & $\color{RedOrange}-0.435$ & $\color{Brown}0.072 \dagger$ \\ 
HC Probability & $\color{YellowOrange}-0.611$ & $\color{Brown}-0.178$ \\ 
Scrubbing Complexity & $\color{YellowOrange}0.696$ & $\color{YellowOrange}0.633$ \\ 
Path Compressibility & $\color{YellowOrange}0.622$ & $\color{Brown}0.230$ \\ 
A* Difficulty & $\color{YellowOrange}0.621$ & $\color{YellowOrange}0.594$ \\ 
Heuristic Error & $\color{YellowOrange}0.629$ & $\color{YellowOrange}0.606$ \\ 
Depression Width & -- & $\color{YellowOrange}0.737$ \\ 
Depression Capacity & $\color{YellowOrange}0.737$ & -- \\
\end{tabular}
\caption{Correlation among complexity measures.} \label{tab:corrComp}
\vspace{-0.5cm}
\end{table}

We observe that HC region size is fairly correlated to HC probability ($\rho = 0.759$) and path compressibility ($\rho = 0.697$).  Likewise, HC probability and path compressibility are highly correlated ($\rho = 0.920$).  These three complexity measures are all directly dependent on the movement of a hill-climbing agent through the search space.

There are  differences in what these complexity measures are capturing.  Unlike the other two measures, HC region size is measuring a space, rather than along a single path.  HC region size differs more appreciably from the other two measures in search spaces with many narrow corridors, versus search spaces with open regions (Section~\ref{chap:combin}).  Additionally, we hypothesize that HC region size is a locally focused measure of hill-climbability among local groups of states, whereas HC probability measures hill-climbability between states that are arbitrarily positioned across the search space.  HC probability provides a qualitative measure of hill-climbing performance, whereas path compressibility provides a quantitative measure.  The former only distinguishes between problems which a hill-climbing agent can and cannot solve, whereas path compressibility provides a gauge of how many failures a hill-climbing agent would encounter.

Heuristic error exhibited a strong correlation to scrubbing complexity ($\rho = 0.945$) and A* difficulty ($\rho = 0.983$).  This is unsurprising, since higher magnitude heuristic errors will naturally correspond to a larger number of states being entered or considered by an LRTA* or A* agent.  A* difficulty and scrubbing complexity are also very highly correlated ($\rho = 0.918$).  We suspect this is due to their mutual sensitivity to inaccurate heuristics.


\subsection{Algorithms versus Complexity Measures}

For each pair of one of the five algorithms and one of the eight complexity measures, we compute the two sets of correlation coefficients using $200$ data points (one for each sub-map). The closer the correlation coefficient is to $+1$ or $-1$, the higher the association of the algorithm's performance to the complexity measure. The complete sets of correlation coefficients are presented in Tables ~\ref{tab:corrMean} and~\ref{tab:corrMedian}, and discussed below, grouped by algorithm.


\begin{table}[htbp]
\center
\begin{tabular}{ l | c | c | c }
& LRTA* & D LRTA* & kNN LRTA* \\ \hline \hline
HC Region Size & $\color{Brown}0.045 \dagger$ & $\color{Brown}-0.207$ & $\mathbf{\color{OliveGreen}-0.804}$ \\ 
HC Probability & $\color{RedOrange}-0.433$ & $\color{RedOrange}-0.309$ & $\color{YellowOrange}-0.708$ \\ 
Scrubbing Complexity & $\mathbf{\color{OliveGreen}0.955}$ & $\mathbf{\color{RedOrange}0.451}$ & $\color{Brown}0.216$ \\
Path Compressibility & $\color{RedOrange}0.498$ & $\color{RedOrange}0.327$ & $\color{YellowOrange}0.617$ \\
A* Difficulty & $\color{OliveGreen}0.854$ & $\color{RedOrange}0.328$ & $\color{Brown}0.136 \dagger$ \\
Heuristic Error & $\color{OliveGreen}0.884$ & $\color{RedOrange}0.352$ & $\color{Brown}0.152$ \\
Depression Width & $\color{YellowOrange}0.663$ & $\color{RedOrange}0.447$ & $\color{YellowOrange}0.503$ \\
Depression Capacity & $\color{YellowOrange}0.647$ & $\color{RedOrange}0.359$ & $\color{Brown}0.066 \dagger$ \\
\end{tabular}

\vspace{0.4cm}

\begin{tabular}{ l | c | c }
& HCDPS & TBA* \\ \hline \hline
HC Region Size & $\color{RedOrange}-0.458$ & $-\color{Brown}0.124 \dagger$ \\ 
HC Probability & $\mathbf{\color{YellowOrange}-0.515}$ & $\color{YellowOrange}-0.513$ \\ 
Scrubbing Complexity & $\color{Brown}0.168$ & $\color{OliveGreen}0.878$ \\ 
Path Compressibility & $\color{RedOrange}0.428$ & $\color{YellowOrange}0.572$ \\ 
A* Difficulty & $\color{Brown}0.108 \dagger$ & $\mathbf{\color{OliveGreen}0.883}$ \\ 
Heuristic Error & $\color{Brown}0.128 \dagger$ & $\color{OliveGreen}0.882$ \\ 
Depression Width & $\color{RedOrange}0.332$ & $\color{YellowOrange}0.705$ \\ 
Depression Capacity & $\color{Brown}0.190 \dagger$ & $\color{YellowOrange}0.613$ \\
\end{tabular}
\caption{Correlation between mean suboptimality and complexity measures.  Highest correlation for each algorithm is in bold.} \label{tab:corrMean}
\vspace{-0.5cm}

\end{table}

\begin{table}[htbp]
\center
\begin{tabular}{ l | c | c | c }
& LRTA* & D LRTA* & kNN LRTA* \\ \hline \hline
HC Region Size & $\color{YellowOrange}-0.609$ & $\color{Brown}-0.211$ & $\color{OliveGreen}-0.762$ \\ 
HC Probability & $\color{OliveGreen}-0.841$ & $\color{Brown}0.073 \dagger$ & $\mathbf{\color{OliveGreen}-0.832}$ \\ 
Scrubbing Complexity & $\color{YellowOrange}0.669$ & $\color{RedOrange}-0.377$ & $\color{RedOrange}0.320$ \\
Path Compressibility & $\mathbf{\color{OliveGreen}0.852}$ & $\color{Brown}-0.071 \dagger$ & $\color{YellowOrange}0.727$ \\
A* Difficulty & $\color{YellowOrange}0.656$ & $\color{RedOrange}-0.429$ & $\color{RedOrange}0.252$ \\
Heuristic Error & $\color{YellowOrange}0.656$ & $\mathbf{\color{RedOrange}-0.443}$ & $\color{RedOrange}0.278$ \\
Depression Width & $\color{YellowOrange}0.728$ & $\color{Brown}-0.127 \dagger$ & $\color{YellowOrange}0.504$ \\
Depression Capacity & $\color{RedOrange}0.352$ & $\color{Brown}-0.248$ & $\color{Brown}0.046 \dagger$ \\
\end{tabular}

\vspace{0.4cm}

\begin{tabular}{ l | c | c }
& HCDPS & TBA* \\ \hline \hline
HC Region Size & $\color{YellowOrange}-0.639$ & $\color{Brown}-0.176$ \\ 
HC Probability & $\mathbf{\color{OliveGreen}-0.828}$ & $\color{YellowOrange}-0.609$ \\ 
Scrubbing Complexity & $\color{RedOrange}0.485$ & $\color{OliveGreen}0.862$ \\ 
Path Compressibility & $\color{OliveGreen}0.809$ & $\color{YellowOrange}0.665$ \\ 
A* Difficulty & $\color{RedOrange}0.449$ & $\color{OliveGreen}0.889$ \\ 
Heuristic Error & $\color{RedOrange}0.467$ & $\mathbf{\color{OliveGreen}0.892}$ \\ 
Depression Width & $\color{YellowOrange}0.548$ & $\color{YellowOrange}0.651$ \\ 
Depression Capacity & $\color{Brown}0.170$ & $\color{YellowOrange}0.542$ \\
\end{tabular}
\caption{Correlation between median suboptimality and complexity measures.  Highest correlation for each algorithm is in bold.} \label{tab:corrMedian}
\vspace{-0.3cm}
\end{table}


\medskip\subsubsection{LRTA*}

We observed a high correlation between the mean suboptimality of LRTA* and scrubbing complexity ($\rho_\text{mean}=0.955$).  This is intuitive, since scrubbing complexity is directly derived from LRTA*.  A high correlation to heuristic error is also observed ($\rho_\text{mean}=0.884$), which we attribute to the link between high magnitude heuristic errors and repeated state revisitation in LRTA*.  The moderate correlation to depression width ($\rho_\text{mean}=0.663$) and depression capacity ($\rho_\text{mean}=0.647$) fits with prior literature that links the presence of heuristic depressions to poor LRTA* performance~\cite{DBLP:journals/jair/BulitkoL06}.

When we consider median suboptimality, LRTA* exhibits higher correlations to the HC-related measures path compressibility ($\rho_\text{median}=0.852$) and HC probability ($\rho_\text{mean}=-0.841$).  In the median case, LRTA* is not as hampered by scrubbing.  By removing outliers, we are removing the cases where LRTA* must perform excessive state revisitation.  We believe that this similarity in behaviour of LRTA* and a hill-climbing agent on easier search problems causes these higher correlations in the median case.


\medskip\subsubsection{D LRTA*}

Despite having the same underlying agent as LRTA* and kNN LRTA*, D LRTA* exhibits no strong correlations with the presented complexity measures.  The interaction between the clique abstraction and the heuristic errors on the map can be complex, even among states within a common partition.  Very suboptimal solutions are usually tied to scrubbing behavior within a partition.  However, the frequency of these cases is only weakly linked to overall scrubbing complexity ($\rho_\text{mean}=0.451)$ and to heuristic error ($\rho_\text{mean}=0.352$). Finding a computationally efficient predictor of D LRTA* performance remains an open research goal.


\medskip\subsubsection{kNN LRTA*}

In the mean case, kNN LRTA* performance is most correlated to HC region size ($\rho_\text{mean}=-0.804$) and HC probability ($\rho_\text{mean}=-0.708$).  Since database records can only be used when they are HC-reachable relative to the start and goal states, a lower HC probability results in a lower chance of finding an appropriate record, causing search to fall back on LRTA* and therefore yielding a higher suboptimality.  HC region size is suspected to be a marginally stronger predictor of record availability than HC probability since it is a more locally focused measure than HC probability.  Since only the $M$ most similar kNN LRTA* database records are considered for use, localized hill-climbability will be more related to a database record being available.

In the median case, similar correlations are observed to HC probability ($\rho_\text{median}=-0.832$) and HC region size ($\rho_\text{median}=-0.762$).  Path compressibility also has a somewhat higher correlation ($\rho_\text{median}=0.727$).


\medskip\subsubsection{HCDPS}

HCDPS performance is most correlated to HC probability ($\rho_\text{mean}=-0.515$) and the other HC-based measures.  We were initially surprised that HC region size did not correlate more highly to the performance of HCDPS in the mean case, since HC region size is computed using the same abstraction method as in HCDPS databases.  However, it appears that in the mean case, HC region size has a two-fold relationship with solution suboptimality for HCDPS.  Larger HC regions typically lead to lower suboptimality.  This is due to the method that HCDPS uses to pre-compute subgoal records.  When passing through fewer abstract regions, as is expected when region sizes are larger, the database record will be generated using fewer constituent paths, and is expected to be closer to optimal.  However, if HC regions are too large, suboptimality increases as the agent is forced to deviate from the optimal path to pass through the seed states.

In the median case, HCDPS correlates most strongly to HC probability ($\rho_\text{median}=-0.828$) and path compressibility ($\rho_\text{median}=0.809$), and correlates more highly with HC region size than in the mean case ($\rho_\text{median}=-0.639$).  We take this as evidence that in typical cases larger HC regions result in lower suboptimality, while in highly suboptimal cases, larger HC regions can cause increased suboptimality, matching the effects described above.

HCDPS performance is poorly correlated to depression capacity ($\rho_\text{mean}=0.190$, $\rho_\text{median}=0.170$), scrubbing complexity ($\rho_\text{mean}=0.168$, $\rho_\text{median}=0.485$) and heuristic error ($\rho_\text{mean}=0.128$, $\rho_\text{median}=0.467$).  Since HCDPS does not perform heuristic updates, there is never a need to revisit states to fill in heuristic depressions.  Therefore, complexity measures that gauge the magnitude of heuristic inaccuracy and state-revisitation are not correlated to HCDPS performance.


\medskip\subsubsection{TBA*}

The mean suboptimality of TBA* is most highly correlated to A* difficulty ($\rho_\text{mean}=0.883$).  This follows from the dependence of TBA* on a bounded A* agent.  The more node expansions required by the A* agent, the more execution phases that will occur with TBA* following a potentially incomplete path.  We attribute the similarly high correlations of TBA* to heuristic error ($\rho_\text{mean}=0.892$) and scrubbing complexity ($\rho_\text{mean}=0.862$) to the high correlation between these two measures and A* difficulty.  TBA* also has a moderately high correlation to depression width ($\rho_\text{mean}=0.705$).

In the median case, TBA* remains highly correlated to A* difficulty ($\rho_\text{median}=0.889$), heuristic error ($\rho_\text{median}=0.892$) and scrubbing complexity ($\rho_\text{median}=0.862$).  This leads us to believe that there are no drastic differences between typical and degenerate behaviour of TBA* relative to the eight complexity measures we present.


\subsection{Correlation to Database Construction Time}

The amount of time required for database precomputation can differ substantially even among search spaces of the same size.  Therefore, we also examined how the complexity measures correlate to precomputation time.  Table~\ref{tab:corrTime} presents the Spearman correlation $\rho_\text{time}$ between complexity measure values and the mean database precomputation time.

\begin{table}[htbp]
\center
\begin{tabular}{ l | c | c | c }
& D LRTA* & kNN LRTA* & HCDPS \\ \hline \hline
HC Region Size & $\color{YellowOrange}-0.544$ & $\color{Brown}-0.016 \dagger$ & $\color{RedOrange}0.477$ \\ 
HC Probability & $\color{YellowOrange}-0.632$ & $\color{RedOrange}-0.491$ & $\mathbf{\color{YellowOrange}0.708}$ \\ 
Scrubbing Complexity & $\color{YellowOrange}0.509$ & $\color{OliveGreen}0.911$ & $\color{RedOrange}-0.344$ \\ 
Path Compressibility & $\mathbf{\color{YellowOrange}0.723}$ & $\color{YellowOrange}0.580$ & $\color{YellowOrange}-0.665$ \\ 
A* Difficulty & $\color{YellowOrange}0.582$ & $\mathbf{\color{OliveGreen}0.988}$ & $\color{RedOrange}-0.342$ \\ 
Heuristic Error & $\color{YellowOrange}0.556$ & $\color{OliveGreen}0.982$ & $\color{RedOrange}-0.357$ \\ 
Depression Width & $\color{RedOrange}0.427$ & $\color{YellowOrange}0.601$ & $\color{RedOrange}-0.444$ \\ 
Depression Capacity & $\color{Brown}0.188$ & $\color{YellowOrange}0.606$ & $\color{Brown}-0.177$ \\
\end{tabular}
\caption{Correlation between complexity measures and database computation times.} \label{tab:corrTime}
\vspace{-0.4cm}
\end{table}

D LRTA* construction time is most correlated to path compressibility ($\rho_\text{time}=0.723$). kNN LRTA* is highly correlated to A* difficulty ($\rho_\text{time}=0.988$).  The most time consuming component of kNN LRTA* database construction requires the optimal solving of a fixed number of search problems with A*.  Therefore, the more difficult these problems are to solve for A*, the longer the time required to build the database.  HCDPS is most correlated to HC probability ($\rho_\text{time}=0.708$).  A high HC probability will result in an abstraction containing fewer HC regions.  This in turn requires the database to compute fewer paths, resulting in a lower precomputation time.


\section{Predictive Modelling}
\label{chap:predictive}

In the preceding section, we established a statistical link between the complexity measures and real-time search performance.  Given the values of the complexity measures for two search spaces, we have an idea of how well a real-time heuristic search algorithm will perform in one search space relative to the other.  This information improves our understanding of what search space features the algorithms are sensitive to, and is useful for characterizing search benchmarks.  However, we would also like to apply the complexity measures to assist in algorithm selection and parameter tuning.  To this end, we now adopt a predictive approach to modelling search performance with the complexity measures.

\subsection{Experimental Design}

\begin{figure}[htbp]
\centering
\includegraphics[width=\columnwidth]{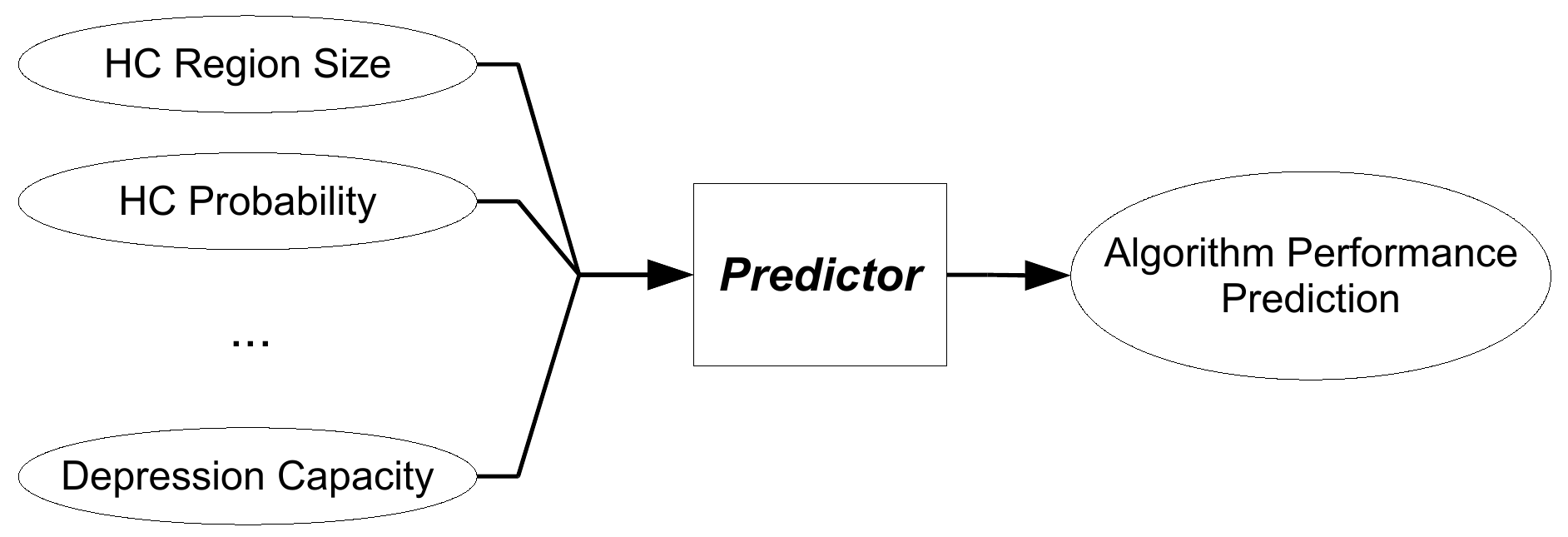}
\caption{Predictor of search performance.} \label{fig:classify}
\vspace{-0.4cm}
\end{figure}

Using machine learning, we construct a predictor of search performance.  As input, this predictor takes the values of the complexity measures computed for a given search space.  As output, the predictor returns a predicted value of a chosen performance metric such as solution suboptimality (Figure~\ref{fig:classify}).

For each metric of search performance, we build two classes of predictors.  For the first class of predictors, we discretize the output metric into $10$ bins with equal frequency in the training data.  The output prediction is then a classification into one of these $10$ bins.  For the second class of predictors, the output is simply a numerical prediction of the search performance.

For each class of predictors, we tested a selection of classifiers within WEKA~\cite{weka}.  We use the naive \emph{ZeroR} classifier as a baseline predictor.  In the first class, \emph{ZeroR} always outputs the first bin.  In the second class, \emph{ZeroR} always outputs the mean value of the output metric observed in the training data.

The predictors are trained and tested on the video game pathfinding data described in Section~\ref{sec:mapExpDesign}.  All trials are performed with $10$-fold cross-validation.  For the first class of predictors, we report the percentage of cases where the correct bin was predicted for the output search metric.  A greater percentage represents higher classifier accuracy.  For the second class of predictors, we report the root-mean-square error (RMSE) and the relative root-square error (RRSE) of the output metric.  RMSE is defined as $\sqrt{\sum_{i=1}^n (\hat{x}_i - x_i)^2 / n}$ where $\hat{x}_i$ and $x_i$ are the predicted and actual values respectively for the $i^th$ datapoint.  RRSE is the percentage size of the error relative to the error of the \emph{ZeroR} classifier.  A smaller RMSE and RRSE correspond to higher classifier accuracy.


\begin{table*}[htbp]\small
\center
\begin{tabular}{ l | c | c | c | c | c }
Classifier & LRTA* & D LRTA* & kNN LRTA* & HCDPS & TBA \\ \hline \hline
BayesNet & $48$ & $18.5$ & $28.5$ & $15.5$ & $32$ \\
MultilayerPerceptron & $50.5$ & $\mathbf{24}$ & $\mathbf{37}$ & $19.5$ & $\mathbf{39.5}$ \\
AdaBoostM1 & $19.5$ & $18.5$ & $18$ & $17$ & $20$ \\
Bagging & $49$ & $22$ & $35.5$ & $16$ & $35.5$ \\
ClassificationViaRegression & $\mathbf{54}$ & $21$ & $31.5$ & $\mathbf{22.5}$ & $33$ \\
RandomCommittee & $51.5$ & $23$ & $29.5$ & $14.5$ & $\mathbf{39.5}$ \\
DecisionTable & $45$ & $18.5$ & $30.5$ & $17.5$ & $31.5$ \\
J48 & $44.5$ & $21$ & $28.5$ & $18$ & $32.5$ \\ \hline
ZeroR & $10$ & $10$ & $10$ & $10$ & $10$ \\
\end{tabular}
\caption{Classification accuracy for discretized mean suboptimality.  Most accurate classifier for each algorithm is  in bold.} \label{tab:predBinMean}
\vspace{-0.4cm}
\end{table*}

\begin{table*}[htbp]\small
\center
\begin{tabular}{ l | c | c | c | c | c }
Classifier & LRTA* & D LRTA* & kNN LRTA* & HCDPS & TBA* \\ \hline \hline
SimpleLinearRegression & $\mathbf{8.975}$, \small{$\mathit{36.22}$} & $\mathbf{6.273}$, \small{$\mathit{76.24}$} & $0.1452$, \small{$\mathit{91.49}$} & $0.0335$, \small{$\mathit{83.48}$} & $0.5925$, \small{$\mathit{58.05}$} \\
LeastMedSq & $23.45$, \small{$\mathit{57.57}$} & $8.555$, \small{$\mathit{103.9}$} & $\mathbf{0.1399}$, \small{$\mathit{88.12}$} & $0.0317$, \small{$\mathit{79.11}$} & $0.4973$, \small{$\mathit{48.72}$} \\
LinearRegression & $13.21$, \small{$\mathit{32.43}$} & $6.563$, \small{$\mathit{79.77}$} & $0.1412$, \small{$\mathit{88.99}$} & $\mathbf{0.0317}$, \small{$\mathit{78.98}$} & $\mathbf{0.4631}$, \small{$\mathit{45.37}$} \\
MultilayerPerceptron & $17.55$, \small{$\mathit{43.07}$} & $13.70$, \small{$\mathit{166.5}$} & $0.1706$, \small{$\mathit{107.5}$} & $0.0350$, \small{$\mathit{87.34}$} & $0.5531$, \small{$\mathit{54.18}$}\\ \hline
ZeroR & $40.74$, \small{$\mathit{100}$} & $8.230$, \small{$\mathit{100}$} & $0.1587$, \small{$\mathit{100}$} & $0.0401$, \small{$\mathit{100}$} & $1.021$, \small{$\mathit{100}$} \\
\end{tabular}
\caption{Prediction error (RMSE) for raw mean suboptimality.  RRSE (\%) is in italics.  Most accurate predictor is  in bold.} \label{tab:predMean}
\vspace{-0.4cm}
\end{table*}

\subsection{Predicting Mean Suboptimality}

Table~\ref{tab:predBinMean} presents the accuracy when predicting discretized mean suboptimality.  We observe that predictors for LRTA* ($54\%$), TBA* ($39.5\%$) and kNN LRTA* ($37\%$) have the highest peak classification accuracies.  This is in keeping with our observations in Table~\ref{tab:corrMean}, where these three algorithms had the highest observed correlations to the complexity measures.  Conversely, D LRTA* ($24\%$) and HCDPS ($22.5\%$) have lower peak accuracies.  However, in all cases, we are able to achieve a higher accuracy than the uninformed \emph{ZeroR} classifier ($10\%$).

Table~\ref{tab:predMean} presents the RMSE and RRSE when predicting continuous mean suboptimality.  We observe the lowest minimum error rates for LRTA* (RRSE $= 36.22\%$) and TBA* (RRSE $= 45.37\%$).  kNN LRTA* (RRSE $= 88.12\%$) is not as successfuly predicted as in the discretized case.


\subsection{Predicting Median Suboptimality}

\begin{table*}[htbp]\small
\center
\begin{tabular}{ l | c | c | c | c | c }
Classifier & LRTA* & D LRTA* & kNN LRTA* & HCDPS & TBA \\ \hline \hline
BayesNet & $43$ & $15.5$ & $41.5$ & $29$ & $31$ \\
MultilayerPerceptron & $48$ & $21$ & $\mathbf{47}$ & $31$ & $35$ \\
AdaBoostM1 & $20$ & $17.5$ & $20$ & $19.5$ & $19$ \\
Bagging & $47.5$ & $\mathbf{23.5}$ & $42$ & $\mathbf{35.5}$ & $34$ \\
ClassificationViaRegression & $48$ & $15$ & $45$ & $30.5$ & $36.5$ \\
RandomCommittee & $\mathbf{49}$ & $21.5$ & $38.5$ & $31$ & $37$ \\
DecisionTable & $39.5$ & $15$ & $36.5$ & $27.5$ & $29$ \\
J48 & $48.5$ & $15.5$ & $41$ & $30$ & $\mathbf{38.5}$ \\ \hline
ZeroR & $10$ & $10$ & $10$ & $10$ & $10$ \\
\end{tabular}
\caption{Classification accuracy for discretized median suboptimality.  Most accurate classifier for each algorithm is  in bold.} \label{tab:predBinMedian}
\vspace{-0.4cm}
\end{table*}

\begin{table*}[htbp]\small
\center
\begin{tabular}{ l | c | c | c | c | c }
Classifier & LRTA* & D LRTA* & kNN LRTA* & HCDPS & TBA* \\ \hline \hline
SimpleLinearRegression & $6.994$, \small{$\mathit{92.81}$} & $0.1313$, \small{$\mathit{94.67}$} & $0.0285$, \small{$\mathit{50.29}$} & $0.0175$, \small{$\mathit{53.57}$} & $0.6409$, \small{$\mathit{53.46}$} \\
LeastMedSq & $7.524$, \small{$\mathit{99.86}$} & $0.1283$, \small{$\mathit{91.81}$} & $0.0240$, \small{$\mathit{42.31}$} & $0.0176$, \small{$\mathit{54.03}$} & $0.5198$, \small{$\mathit{43.35}$} \\
LinearRegression & $\mathbf{6.564}$, \small{$\mathit{87.11}$} & $\mathbf{0.1257}$, \small{$\mathit{89.98}$} & $0.0227$, \small{$\mathit{40.13}$} & $\mathbf{0.0173}$, \small{$\mathit{52.86}$} & $\mathbf{0.5194}$, \small{$\mathit{43.33}$} \\
MultilayerPerceptron & $9.149$, \small{$\mathit{121.4}$} & $0.1431$, \small{$\mathit{102.4}$} & $\mathbf{0.0226}$, \small{$\mathit{39.88}$} & $0.0238$, \small{$\mathit{72.90}$} & $0.6725$, \small{$\mathit{56.09}$}  \\ \hline
ZeroR & $7.535$, \small{$\mathit{100}$} & $0.1397$, \small{$\mathit{100}$} & $0.0566$, \small{$\mathit{100}$} & $0.0326$, \small{$\mathit{100}$} & $1.199$, \small{$\mathit{100}$} \\
\end{tabular}
\caption{Prediction error (RMSE) for raw median suboptimality.  RRSE is in italics.  Most accurate predictor is in bold.} \label{tab:predMedian}
\vspace{-0.4cm}
\end{table*}

Table~\ref{tab:predBinMedian} presents the accuracy when predicting discretized median suboptimality.  We again observe that LRTA* ($49\%$), kNN LRTA* ($47\%$) and TBA* ($38.5\%$) have the highest peak classification accuracies.  In the case of kNN LRTA* and HCDPS ($35.5\%$), we observe a higher classification accuracy than in the mean case.  This corresponds to the higher correlations observed with median suboptimality for these algorithms, as presented in Table~\ref{tab:corrMedian}.

Table~\ref{tab:predMedian} presents the RMSE and RRSE when predicting continuous median suboptimality.  We observe the lowest minimum error rates for kNN LRTA* (RRSE $= 39.88\%$), TBA* (RRSE $= 43.33\%$) and HCDPS (RRSE $= 52.86\%$).


\begin{table*}[htbp]\small
\center
\begin{tabular}{ l | c | c | c }
Classifier & D LRTA* & kNN LRTA* & HCDPS \\ \hline \hline
BayesNet & $21$ & $67.5$ & $18.5$ \\
MultilayerPerceptron & $\mathbf{32}$ & $63$ & $21.5$ \\
AdaBoostM1 & $19$ & $19.5$ & $18.5$ \\
Bagging & $31$ & $\mathbf{69}$ & $\mathbf{25.5}$ \\
ClassificationViaRegression & $29$ & $61$ & $24$ \\
RandomCommittee & $27.5$ & $66$ & $24$ \\
DecisionTable & $20$ & $62.5$ & $18.5$\\
J48 & $27.5$ & $62$ & $22$ \\ \hline
ZeroR & $10$ & $10$ & $10$ \\
\end{tabular}
\caption{Classification accuracy for discretized database construction time.  Most accurate classifier for each algorithm is  in bold.} \label{tab:predBinBuild}
\vspace{-0.4cm}
\end{table*}

\begin{table*}[htbp]\small
\center
\begin{tabular}{ l | c | c | c }
Classifier & D LRTA* & kNN LRTA* & HCDPS \\ \hline \hline
SimpleLinearRegression & $12.97$, \small{$\mathit{69.46}$} & $0.3369$, \small{$\mathit{16.00}$} & $0.5362$, \small{$\mathit{75.49}$} \\
LeastMedSq & $13.24$, \small{$\mathit{70.90}$} & $0.3007$, \small{$\mathit{14.29}$} & $0.5346$, \small{$\mathit{75.26}$} \\
LinearRegression & $\mathbf{11.19}$, \small{$\mathit{59.94}$} & $\mathbf{0.2672}$, \small{$\mathit{12.70}$} & $\mathbf{0.5113}$, \small{$\mathit{71.99}$} \\
MultilayerPerceptron & $12.64$, \small{$\mathit{67.71}$} & $0.3580$, \small{$\mathit{17.01}$} & $0.6813$, \small{$\mathit{95.92}$} \\ \hline
ZeroR & $18.68$, \small{$\mathit{100}$} & $2.105$, \small{$\mathit{100}$} & $0.7103$, \small{$\mathit{100}$} \\
\end{tabular}
\caption{Prediction error (RMSE) for raw database construction time.  RRSE is in italics.  Most accurate predictor is in bold.} \label{tab:predBuild}
\vspace{-0.4cm}
\end{table*}


\subsection{Predicting Database Construction Time}

Table~\ref{tab:predBinBuild} presents the accuracy when predicting discretized database construction time.  We observe the highest peak classification accuracy for kNN LRTA* ($69\%$).  In Table~\ref{tab:corrTime}, kNN LRTA* database construction time had the highest observed correlations.  Note though that D LRTA* ($32\%$) and HCDPS ($25.5\%$) also exhibit peak classification accuracies well above the \emph{ZeroR} classifier.

Table~\ref{tab:predBuild} presents the RMSE and RRSE when predicting raw database construction time.  Similarly to in the discrete case, the error rate for kNN LRTA* ($12.70\%$) is lowest, followed by D LRTA* ($59.94\%$) and then HCDPS ($71.99\%$).

\begin{figure}[htbp]
\centering
\includegraphics[width=\columnwidth]{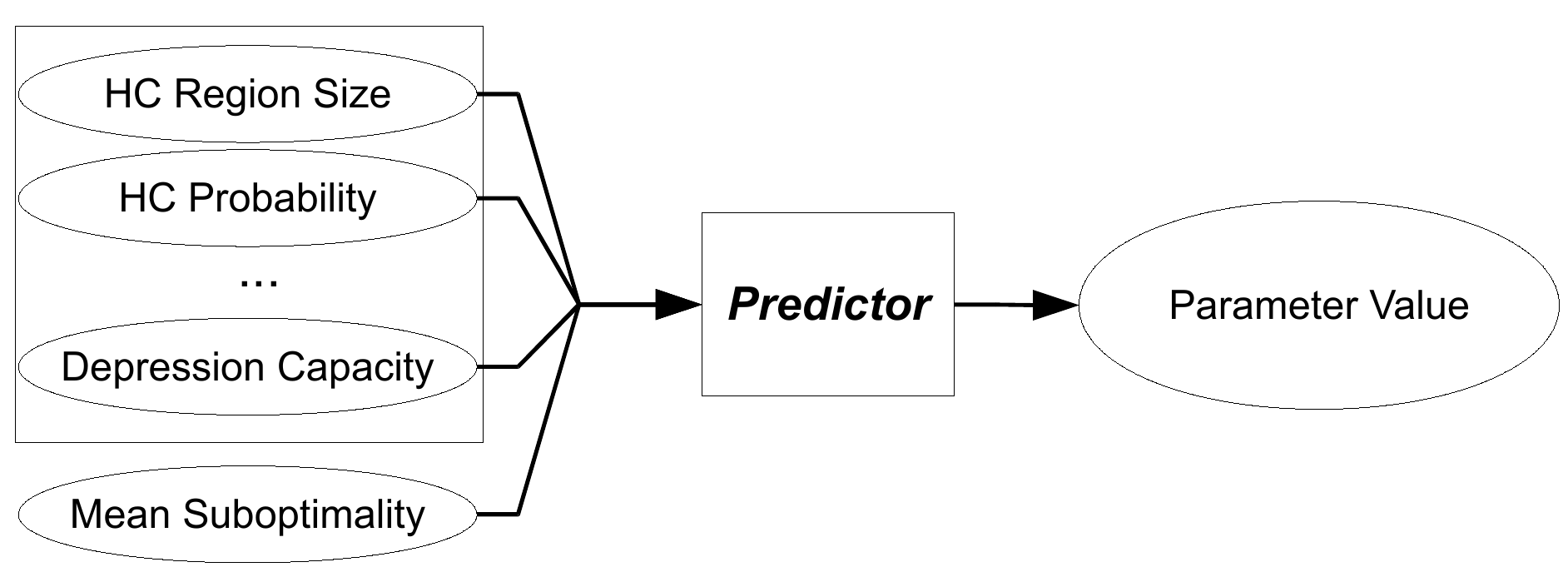}
\caption{Predictor for assisting algorithm parameterization.} \label{fig:classify2}
\vspace{-0.4cm}
\end{figure}


\subsection{Assisting Algorithm Parameterization}

In the previous section we use the values of complexity measures to predict the search performance of a given algorithm for a search space.  However, if we instead include \emph{desired} search performance as an input to the predictor, we can predict appropriate values for algorithm parameters (Figure~\ref{fig:classify2}).

We use this approach to predict appropriate kNN LRTA* database sizes that will achieve a desired mean suboptimality.  Using the same search problems described in Section~\ref{sec:mapExpDesign}, we ran kNN LRTA* with database sizes of $500$, $1000$, $2500$, $5000$, $10000$ and $20000$, for $6 \times 200 = 1200$ datapoints.  We then machine-learned predictors to output database size, using the values of the eight complexity measures for a map and the mean kNN LRTA* suboptimality for that map as input.

Table~\ref{tab:predBinDBSize} presents the accuracy when predicting database size discretized into six bins.  Table~\ref{tab:predDBSize} presents the RMSE and RRSE when predicting continuous database size.  All experiments are performed with $10$-fold cross-validation.  In the discrete case, the peak classification accuracy is $50.42\%$.  In the continuous case, the lowest RRSE is ($52.54\%$).  In both cases, we are able to outperform the \emph{ZeroR} classifier.

\begin{table}[htbp]\small
\center
\begin{tabular}{ l | c }
Classifier & Accuracy \\ \hline \hline
BayesNet & $34.17\%$ \\
MultilayerPerceptron & $49.33\%$ \\
AdaBoostM1 & $30.33\%$ \\
Bagging & $\mathbf{50.42\%}$ \\
ClassificationViaRegression & $48.58\%$ \\
RandomCommittee & $39.83\%$ \\
DecisionTable & $39.12\%$ \\
J48 & $48.92\%$ \\ \hline
ZeroR & $16.67\%$ \\
\end{tabular}
\caption{Classification accuracy for kNN LRTA* database size.  Most accurate classifier is in bold.} \label{tab:predBinDBSize}
\vspace{-0.4cm}
\end{table}

\begin{table}[htbp]\small
\center
\begin{tabular}{ l | c }
Classifier & RMSE, \small{\textit{(RRSE \%)}} \\ \hline \hline
SimpleLinearRegression & $6469.66$, \small{$\mathit{94.79}$} \\
Bagging & $\mathbf{3585.58}$, \small{$\mathit{52.54}$} \\
LinearRegression & $6840.78$, \small{$\mathit{94.96}$} \\
MultilayerPerceptron & $4130.0811$, \small{$\mathit{51.76}$} \\ \hline
ZeroR & $6824.65$, \small{$\mathit{100}$} \\
\end{tabular}
\caption{Prediction error (RMSE) for kNN LRTA* database size.  RRSE is in italics.  Most accurate predictor is in bold.} \label{tab:predDBSize}
\vspace{-0.4cm}
\end{table}


\section{Beyond Video-game Pathfinding}
\label{chap:combin}

In the previous two sections we presented a set of complexity measures for characterizing search spaces, and demonstrated their relationship to the performance of five real-time heuristic search algorithms in the domain of video-game pathfinding.  In this section, we seek to extend these results to two additional classes of search problems: pathfinding in mazes and road maps.  This is the first study to apply TBA*, kNN LRTA* and HCDPS to either of these domains, or any search space outside of video-game pathfinding.  

We begin by  describing the two new classes of search problems.  We then conduct experiments using the same correlation approach as in the previous section.  To mitigate the influence of differing search space sizes, we again apply the sampling method described in Section~\ref{sec:mapExpDesign}.  All of the experiments are performed using sub-spaces of $20$ thousand states.

For this section we omit the algorithm D LRTA*.  This decision is in part due to the poor correlations observed to D LRTA* in the previous section.  Additionally, computing a D LRTA* database becomes exceedingly expensive because finding cliques in general graphs is an expensive task.


\subsection{Road Maps}

Finding short paths in real-time on road maps is a common demand for consumer electronics such as personal and automotive GPS devices.  Similar to video-game pathfinding, this is an application where speed of search and quality of paths are critical to a positive user experience.  The edge weights in the graph are real-valued and represent the geographical distance between the states the edge connects.  For each state, we also have integer longitude and latitude values that are used for heuristic computation.

We use four maps from the Ninth DIMACS Implementation Challenge~\cite{dimacs:roadmaps}.  The road maps have $0.26$ to $1.07$ million states and $0.73$ to $2.71$ million edges.  For each of the four maps, we generated $25$ sub-spaces for a total of $100$ submaps.  We solved $250$ search problems on each map.  We use the geographical Euclidean distance between two nodes as our heuristic.  For states $s_1$ and $s_2$, $h(s_1,s_2) = \sqrt{(x_1 - x_2)^2 + (y_1 - y_2)^2} $ where $x_i$ and $y_i$ are the longitude and latitude of state $s_i$ respectively.  Each search problem has a single start state and a single goal state chosen randomly from the search space.

Note that HCDPS is omitted from this section.  Constructing an HCDPS database for these road maps exceeded the available amount of memory.  The HCDPS partitioning scheme generated too many abstract regions for road maps.  This caused the number of required subgoal entries to increase beyond a feasible level.  This is reflected in the very low HC region size observed for road maps ($5$ to $15$ states on average).


\begin{table}[htbp]\scriptsize
\center
\begin{tabular}{ l | c | c | c }
& HC Region Size & HC Probability & Scrub. Complexity \\ \hline \hline
HC Region Size & -- & $0.683$ & $\mathbf{-0.569}$ \\ 
HC Probability & $0.683$ & -- & $-0.611$ \\ 
Scrubbing Complexity & $\mathbf{-0.569}$ & $-0.611$ & -- \\
Path Compressibility & $-0.706$ & $-0.764$ & $0.830$ \\
A* Difficulty & $\mathbf{-0.364}$ & $-0.468$ & $0.731$ \\
Heuristic Error & $\mathbf{-0.667}$ & $-0.732$ & $0.892$ \\
Depression Width & $\mathbf{-0.936}$ & $-0.654$ & $0.570$ \\
Depression Capacity & $\mathbf{-0.710}$ & $\mathbf{-0.746}$ & $0.836$ \\
\end{tabular}

\vspace{0.4cm}

\begin{tabular}{ l | c | c | c }
& Path Compressibility & A* Difficulty & Heuristic Error \\ \hline \hline
HC Region Size & $-0.706$ & $\mathbf{-0.364}$ & $\mathbf{-0.667}$ \\ 
HC Probability & $-0.764$ & $-0.468$ & $-0.732$ \\ 
Scrubbing Complexity & $0.830$ & $0.732$ & $0.892$ \\ 
Path Compressibility & -- & $0.653$ & $\mathbf{0.955}$ \\ 
A* Difficulty &  $0.653$ & -- & $0.696$ \\ 
Heuristic Error & $\mathbf{0.955}$ & $0.696$ & -- \\
Depression Width & $0.644$ & $0.454$ & $0.607$ \\
Depression Capacity & $\mathbf{0.924}$ & $0.561$ & $\mathbf{0.954}$ \\
\end{tabular}

\vspace{0.4cm}

\begin{tabular}{ l | c | c }
& Depression Width & Depression Capacity \\ \hline \hline
HC Region Size & $\mathbf{-0.936}$ & $\mathbf{-0.710}$ \\ 
HC Probability & $-0.654$ & $\mathbf{-0.746}$ \\ 
Scrubbing Complexity & $0.570$ & $0.835$ \\ 
Path Compressibility & $0.644$ & $\mathbf{0.924}$ \\ 
A* Difficulty & $0.454$ & $0.561$ \\ 
Heuristic Error & $0.607$ & $\mathbf{0.954}$ \\ 
Depression Width & -- & $0.636$ \\ 
Depression Capacity & $0.636$ & -- \\
\end{tabular}
\caption{Correlation between complexity measures for road maps.  Correlations substantially different from those observed for video-game pathfinding are indicated in bold.} \label{tab:corrRoadComp}
\vspace{-0.4cm}
\end{table}


\subsection{Correlation Among Complexity Measures}

Table~\ref{tab:corrRoadComp} presents the observed Spearman rank correlations between the complexity measures for road maps.  Most of the correlations match those observed for video-game pathfinding, and so we refer back to the rationale provided in Section~\ref{sec:amongComp}.  However, there are several instances where the observed correlation differs by a substantial margin (greater than $\pm0.3$).  In particular, HC region size  and depression capacity exhibited higher correlations in road maps than in video-game maps.

\begin{table}[htbp]
\center
\begin{tabular}{ l | c | c | c }
& LRTA* & kNN LRTA* & TBA* \\ \hline \hline
LRTA* & -- & $0.970$ & $0.189 \dagger$ \\ 
kNN LRTA* & $0.970$ & -- & $0.220$ \\ 
TBA* & $0.189 \dagger$ & $0.220$ & -- \\
\end{tabular}
\caption{Correlation for mean solution suboptimality for road maps.} \label{tab:corrRoadAlgMean}
\vspace{-0.4cm}
\end{table}

\begin{table}[htbp]
\center
\begin{tabular}{ l | c | c | c }
& LRTA* & kNN LRTA* & TBA* \\ \hline \hline
LRTA* & -- & $0.984$ & $0.137 \dagger	$ \\ 
kNN LRTA* & $0.984$ & -- & $0.133 \dagger$ \\ 
TBA* & $0.137 \dagger$ & $0.133 \dagger$ & -- \\
\end{tabular}
\caption{Correlation for median solution suboptimality for road maps.} \label{tab:corrRoadAlgMedian}
\vspace{-0.4cm}
\end{table}


\subsection{Correlation Among Algorithms}
\label{sec:roadMapAlgCorr}

Table~\ref{tab:corrRoadAlgMean} presents the correlation between the mean suboptimality of the algorithms for road maps.  Table~\ref{tab:corrRoadAlgMedian} presents the correlation between the median suboptimality of the algorithms for road maps.  In both cases, a very high correlation is observed between LRTA* and kNN LRTA* suboptimality.  In the vast majority of road map problems, kNN LRTA* is unable to find a suitable database record, and thus falls back on LRTA*.  Therefore, the two algorithms are observed to have very similar performance.  This is reflected in the very low observed HC probability for road maps (less than $0.03$).

\begin{table}[htbp]
\center
\begin{tabular}{ l | c | c | c }
& LRTA* & kNN LRTA* & TBA* \\ \hline \hline
HC Region Size & $-0.431$ & $-0.528$ & $0.071 \dagger$ \\ 
HC Probability & $-0.402$ & $-0.502$ & $0.017 \dagger$ \\ 
Scrubbing Complexity & $\mathbf{0.878}$ & $\mathbf{0.908}$ & $0.051 \dagger$ \\
Path Compressibility & $0.596$ & $0.697$ & $-0.001 \dagger$ \\
A* Difficulty & $0.737$ & $0.773$ & $\mathbf{0.536}$ \\
Heuristic Error & $0.680$ & $0.761$ & $0.009 \dagger$ \\
Depression Width & $0.510$ & $0.592$ & $0.048 \dagger$ \\
Depression Capacity & $0.590$ & $0.689$ & $-0.100 \dagger$ \\
\end{tabular}
\caption{Correlation of mean suboptimality and complexity measures for road maps.  The strongest correlation for each algorithm is in bold.} \label{tab:corrRoadMean}
\vspace{-0.4cm}
\end{table}

\begin{table}[htbp]
\center
\begin{tabular}{ l | c | c | c }
& LRTA* & kNN LRTA* & TBA* \\ \hline \hline
HC Region Size & $-0.680$ & $-0.733$ & $0.133 \dagger$ \\ 
HC Probability & $-0.648$ & $-0.699$ & $0.059 \dagger$ \\ 
Scrubbing Complexity & $\mathbf{0.839}$ & $\mathbf{0.844}$ & $-0.007 \dagger$ \\
Path Compressibility & $0.759$ & $0.812$ & $-0.030 \dagger$ \\
A* Difficulty & $0.739$ & $0.743$ & $\mathbf{0.491}$ \\
Heuristic Error & $0.776$ & $0.818$ & $-0.016 \dagger$ \\
Depression Width & $0.773$ & $0.807$ & $-0.031 \dagger$ \\
Depression Capacity & $0.740$ & $0.783$ & $-0.143 \dagger$ \\
\end{tabular}
\caption{Correlation of median suboptimality and complexity measures for road maps.  The strongest correlation for each algorithm is in bold.} \label{tab:corrRoadMedian}
\vspace{-0.4cm}
\end{table}


\subsection{Correlation Between Complexity and Performance}

Table~\ref{tab:corrRoadMean} presents the correlation between the mean suboptimality of the algorithms and the complexity measure values for road maps.  Table~\ref{tab:corrRoadMedian} presents the correlation between the median suboptimality of the algorithms and the complexity measure values for road maps.  In both the mean and median case, LRTA* suboptimality is correlated to similar measures as in video-game pathfinding, with the highest correlation to scrubbing complexity ($\rho_\text{mean} = 0.878$, $\rho_\text{median} = 0.839$).  The higher correlation between LRTA* and HC region size in road maps ($\rho_\text{mean} = -0.431$, $\rho_\text{median} = -0.680$) parallels the increased correlation between HC region size and scrubbing complexity in Table~\ref{tab:corrRoadComp}.

kNN LRTA* is most correlated to scrubbing complexity ($\rho_\text{mean} = -0.431$, $\rho_\text{median} = -0.680$), A* difficulty ($\rho_\text{mean} = 0.773$, $\rho_\text{median} = 0.743$) and heuristic error ($\rho_\text{mean} = 0.761$, $\rho_\text{median} = 0.818$).  In video-game pathfinding, these measures had a very low correlation to kNN LRTA* suboptimality.  We attribute this difference to the lower likelihood that kNN LRTA* can find a subgoal in road maps, as discussed in Section~\ref{sec:roadMapAlgCorr}.

TBA* suboptimality in road maps is most correlated to A* difficulty ($\rho_\text{mean} = 0.536$, $\rho_\text{median} = 0.491$).  As discussed in the previous section, this is due to the foundation of TBA* being a depth-limited A* agent.


\subsection{Mazes}

Mazes are defined identically to the video-game maps used in section~\ref{chap:videogames}.  However, search problems in a maze are generally more challenging than in a video-game map.  Video-game maps are typically composed of open regions where obstacles may frequently have no impact on direct agent movement.  In contrast, mazes consist of many winding narrow passages, often forcing an agent to take a convoluted path to reach the goal state.  The practical manifestation of this tendency is reduced heuristic accuracy.

Another property that differentiates mazes from video games results from corridor width.  The \emph{corridor width} of a maze is defined as the width in cells of the narrow passages constituting the maze.  The mazes we explore have corridor widths of $1$, $2$, $4$ and $8$.  As a result, the branching factor of states is typically smaller in mazes than in video-game maps.  

\begin{figure}[htbp]
\centering
\includegraphics[width=\columnwidth]{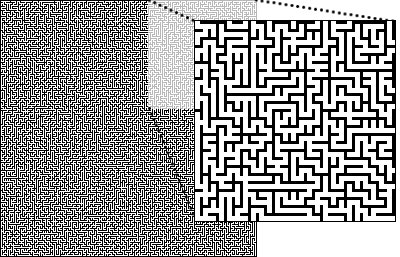}
\caption{An example $512 \times 512$ maze, with a section enlarged for visibility.  This maze has a corridor width of $1$.} \label{fig:mazeFig}
\end{figure}

We use a set of $20$ distinct random mazes~\cite{movingAI}, with $5$ mazes for each corridor width.  The mazes are $512 \times 512$ cells in size.    An example maze is presented in Figure~\ref{fig:mazeFig}.  For each maze we generated $5$ sub-spaces of $20$ thousand states, for a total of $100$ sub-mazes.  We solved $250$ search problems in each sub-maze.


\begin{table}[htbp]\scriptsize
\center
\begin{tabular}{ l | c | c | c }
& HC Region Size & HC Probability & Scrub. Complexity \\ \hline \hline
HC Region Size & -- & $0.899$ & $-0.049 \dagger$ \\ 
HC Probability & $0.899$ & -- & $\mathbf{-0.109} \dagger$ \\ 
Scrubbing Complexity & $-0.049 \dagger$ & $\mathbf{-0.109} \dagger$ & -- \\
Path Compressibility & $-0.924$ & $-0.900$ & $\mathbf{0.206}$ \\
A* Difficulty & $\mathbf{0.420}$ & $\mathbf{0.321}$ & $\mathbf{0.378}$ \\
Heuristic Error & $\mathbf{-0.837}$ & $-0.829$ & $\mathbf{0.337}$ \\
Depression Width & $\mathbf{-0.921}$ & $-0.890$ & $\mathbf{0.065} \dagger$ \\
Depression Capacity	& $\mathbf{0.951}$ &  $\mathbf{0.890}$ & $\mathbf{-0.060} \dagger$ \\
\end{tabular}

\vspace{0.4cm}

\begin{tabular}{ l | c | c | c }
& Path Compressibility & A* Difficulty & Heuristic Error \\ \hline \hline
HC Region Size & $-0.924$ & $\mathbf{0.421}$ & $\mathbf{-0.837}$ \\ 
HC Probability & $-0.900$ & $\mathbf{0.321}$ & $-0.829$ \\ 
Scrubbing Complexity & $\mathbf{0.206}$ & $\mathbf{0.378}$ & $\mathbf{0.337}$ \\ 
Path Compressibility & -- & $-0.209$ & $0.962$ \\ 
A* Difficulty &  $\mathbf{-0.209}$ & -- & $\mathbf{-0.015} \dagger$ \\ 
Heuristic Error & $0.962$ &  $\mathbf{-0.015} \dagger$ & -- \\
Depression Width & $\mathbf{0.905}$ & $\mathbf{-0.372}$ & $0.833$ \\
Depression Capacity & $\mathbf{-0.932}$ & $0.435$ & $\mathbf{-0.840}$ \\
\end{tabular}

\vspace{0.4cm}

\begin{tabular}{ l | c | c }
& Depression Width & Depression Capacity \\ \hline \hline
HC Region Size & $\mathbf{-0.921}$ & $\mathbf{0.951}$ \\ 
HC Probability & $-0.886$ & $\mathbf{0.890}$ \\ 
Scrubbing Complexity & $\mathbf{0.065} \dagger$ & $\mathbf{-0.060} \dagger$ \\ 
Path Compressibility & $0.905$ & $\mathbf{-0.932}$ \\ 
A* Difficulty & $\mathbf{-0.372}$ & $0.435$ \\ 
Heuristic Error & $0.833$ & $\mathbf{-0.840}$ \\ 
Depression Width & -- & $\mathbf{-0.865}$ \\ 
Depression Capacity & $\mathbf{-0.865}$ & -- \\
\end{tabular}
\caption{Correlation among complexity measures for mazes.  Correlations substantially different from those observed for video-game pathfinding are indicated in bold.} \label{tab:corrMazeComp}
\vspace{-0.4cm}
\end{table}

\subsection{Correlation Among Complexity Measures}

Table~\ref{tab:corrMazeComp} presents the observed correlations between the complexity measures for mazes.  The observed correlations for mazes differ more significantly from video-game pathfinding than those for the road maps did.  For example, several of the correlations to HC region size are much higher.  We suspect that this is due to the clustering of data according to corridor width.  Mazes with a common corridor width typically exhibit a similar HC region size.

The observed correlations for scrubbing complexity are lower than for video-game pathfinding.  Scrubbing complexity appears to have a similar spread of values for each different corridor width.
In contrast, other measures are generally more disparate across different corridor widths.  This decreases the observed correlations for scrubbing complexity.

The most interesting result is that the directions of the correlations for A* difficulty and for depression capacity are reversed from those observed in video-game pathfinding.  We attribute this observation to the Yule-Simpson effect.  Despite a positive trend among the data with mazes of a fixed corridor width, the overall data exhibits a negative trend.  This indicates a potential danger of using a correlation-based analysis of the complexity measures without also considering the distribution of the data, especially with diverse search spaces.


\begin{table}[htbp]
\center
\begin{tabular}{ l | c | c }
& LRTA* & kNN LRTA* \\ \hline \hline
LRTA* & -- & $-0.528$ \\ 
kNN LRTA* & $-0.528$ & -- \\ 
HCDPS & $0.277$ & $-0.422$ \\
TBA* & $0.780$ & $-0.810$ \\
\end{tabular}

\vspace{0.4cm}

\begin{tabular}{ l | c | c }
& HCDPS & TBA* \\ \hline \hline
LRTA* & $0.277$ & $0.780$ \\ 
kNN LRTA* & $-0.422$ & $-0.810$ \\ 
HCDPS & -- & $0.549$ \\
TBA* & $0.549$ & -- \\
\end{tabular}
\caption{Correlation for mean  suboptimality for maze pathfinding.} \label{tab:corrMazeAlgMean}
\vspace{-0.4cm}
\end{table}


\begin{table}[htbp]
\center
\begin{tabular}{ l | c | c }
& LRTA* & kNN LRTA* \\ \hline \hline
LRTA* & -- & $-0.522$ \\ 
kNN LRTA* & $-0.522$ & -- \\ 
HCDPS & $0.238$ & $-0.340$ \\
TBA* & $0.716$ & $-0.829$ \\
\end{tabular}

\vspace{0.4cm}

\begin{tabular}{ l | c | c }
& HCDPS & TBA* \\ \hline \hline
LRTA* & $0.238$ & $0.716$ \\ 
kNN LRTA* & $-0.340$ & $-0.829$ \\ 
HCDPS & -- & $0.474$ \\
TBA* & $0.474$ & -- \\
\end{tabular}
\caption{Correlation for median  suboptimality for maze pathfinding.} \label{tab:corrMazeAlgMedian}
\vspace{-0.4cm}
\end{table}

\subsection{Correlation Among Algorithms}

Table~\ref{tab:corrMazeAlgMean} presents the correlation between the mean suboptimality of the algorithms for mazes.  Table~\ref{tab:corrMazeAlgMedian} presents the correlation between the median suboptimality of the algorithms for mazes.  The observed correlations are analogous to those observed in video-game pathfinding, with the exception of kNN LRTA* where the direction of the correlations is reversed.  We again attribute this to the Yule-Simpson effect.  The correlations for mazes of a single fixed corridor width are in line with those observed in video-game maps.


\begin{table}[htbp]
\center
\begin{tabular}{ l | c | c }
& LRTA* & kNN LRTA* \\ \hline \hline
HC Region Size & $0.786$ & $\mathbf{-0.840}$ \\ 
HC Probability & $0.745$ & $-0.807$ \\ 
Scrubbing Complexity & $0.325$ & $0.203$ \\
Path Compressibility & $-0.759$ & $0.810$ \\
A* Difficulty & $0.429$ & $-0.328$ \\
Heuristic Error & $-0.651$ & $0.745$ \\
Depression Width & $\mathbf{-0.790}$ & $0.806$ \\
Depression Capacity & $0.788$ & $-0.814$ \\
\end{tabular}

\vspace{0.4cm}

\begin{tabular}{ l | c | c }
& HCDPS & TBA* \\ \hline \hline
HC Region Size & $0.466$ & $0.922$ \\ 
HC Probability & $0.440$ & $0.898$ \\ 
Scrubbing Complexity & $-0.384$ & $-0.183 \dagger$ \\
Path Compressibility & $-0.551$ & $\mathbf{-0.978}$ \\
A* Difficulty & $0.031 \dagger$ & $0.283$ \\
Heuristic Error & $\mathbf{-0.583}$ & $-0.935$ \\
Depression Width & $-0.456$ & $-0.903$ \\
Depression Capacity & $0.447$ & $0.928$ \\
\end{tabular}
\caption{Correlation of mean  suboptimality and complexity measures for maze pathfinding.  The strongest correlation for each algorithm is in bold.} \label{tab:corrMazeMean}
\vspace{-0.4cm}
\end{table}


\begin{table}[htbp]
\center
\begin{tabular}{ l | c | c }
& LRTA* & kNN LRTA* \\ \hline \hline
HC Region Size & $0.733$ & $\mathbf{-0.847}$ \\ 
HC Probability & $0.660$ & $-0.827$ \\ 
Scrubbing Complexity & $0.312$ & $0.103 \dagger$ \\
Path Compressibility & $-0.651$ & $0.820$ \\
A* Difficulty & $0.597$ & $-0.348$ \\
Heuristic Error & $-0.506$ & $0.754$ \\
Depression Width & $-0.726$ & $0.809$ \\
Depression Capacity & $\mathbf{0.754}$ & $-0.814$ \\
\end{tabular}

\vspace{0.4cm}

\begin{tabular}{ l | c | c }
& HCDPS & TBA* \\ \hline \hline
HC Region Size & $0.390$ & $0.917$ \\ 
HC Probability & $0.382$ & $0.888$ \\ 
Scrubbing Complexity & $-0.344$ & $-0.167 \dagger$ \\
Path Compressibility & $-0.464$ & $\mathbf{-0.970}$ \\
A* Difficulty & $0.069 \dagger$ & $0.303$ \\
Heuristic Error & $\mathbf{-0.489}$ & $-0.925$ \\
Depression Width & $-0.382$ & $-0.899$ \\
Depression Capacity & $0.384$ & $0.925$ \\
\end{tabular}
\caption{Correlation of median  suboptimality and  complexity measures for maze pathfinding.  The strongest correlation for each algorithm is in bold.} \label{tab:corrMazeMedian}
\vspace{-0.4cm}
\end{table}

\subsection{Correlation Between Complexity and Performance}

Table~\ref{tab:corrMazeMean} presents the correlation between the mean suboptimality of the algorithms and the complexity measure values for mazes.  Table~\ref{tab:corrMazeMedian} presents the correlation between the median suboptimality of the algorithms and the complexity measure values for mazes.  We again observe a manifestation of the Yule-Simpson effect: the direction of most correlations for LRTA*, HCDPS and TBA* are reversed from those observed in video-game pathfinding.	

For kNN LRTA*, HC region size ($\rho_\text{mean} = -0.840$, $\rho_\text{median} = -0.847$), HC probability ($\rho_\text{mean} = -0.807$, $\rho_\text{median} = -0.827$) and path compressibility ($\rho_\text{mean} = 0.810$, $\rho_\text{median} = 0.820$) have similarly high correlations to those in video-game maps.  We suspect that this is due to the way that kNN LRTA* provides subgoals during search.  The corridor width does not significantly alter the ability of kNN LRTA* to find  subgoals.

\section{Discussion}

In this section we conduct a discussion of the results presented in this work.  We organize the discussion according to the three major goals of our research presented in Section~\ref{chap:intro}.  We also discuss how the research could be extended to make additional contributions towards these goals.  Finally, we suggest future work in the field of real-time heuristic search.

\subsection{Understanding Algorithm Performance}

The statistical correlations we observe in Sections~\ref{chap:videogames} and~\ref{chap:combin} provide insight into how search space features affect the performance of real-time heuristic search.  While the importance of these features was already known, we now have a way of empirically measuring the degree to which they are affect the performance a specific algorithm.

Leveraging the new results, we propose to enhance kNN LRTA* with a complexity-aware database construction. Specifically, instead of populating its database with random problems, it can focus on areas of space that are more complex according to one of the complexity measures.

\subsection{Facilitating Algorithm Use}

We have described a system by which one can inform the decision of which real-time algorithm is suitable for a given search space.  Using the predictors we presented, one can automatically predict the performance of an algorithm on a novel instance of a search space.  One can also use a similar predictor to determine algorithm parameters which will yield the approximate desired level of performance.  Previously there was no way to accomplish this without expert knowledge.  Given the large investment of time required to compute databases for subgoal-driven search in large search spaces, this may be a valuable tool.

\subsection{Characterizing Benchmarks}

The importance of characterizing benchmarks has been recognized and recent on-line repositories (e.g.,~\cite{movingAI}) have begun listing some complexity measures for the search spaces posted. We support that effort by proposing that the eight complexity measures we have presented are appropriate for characterizing benchmarks for real-time heuristic search.  Having a universal way of empirically measuring the complexity of search spaces is a useful tool for algorithm development and evaluation.  The complexity measures can be applied to any search space, and have been statistically demonstrated to impact the performance of real-time heuristic search.

\subsection{Future Work}

Several improvements could be made to bolster the strength of the predictors we present.  The performance of some algorithms, such as D LRTA*, was not predicted as accurately as others.  We suspect that this could be improved by finding a better way to extract search space features which affect D LRTA* performance.  We tested additional measures to try and measure the effect of the D LRTA* clique abstraction on search performance.  However, these measures were expensive to compute and did not yield significantly stronger results.

Rayner et. al recently examined dimensionality as an intrinsic property of search spaces~\cite{DBLP:conf/aaai/RaynerBS11}.  They found that by considering the dimensionality of a search space, they were able to gain insights as to what classes of heuristic function would be appropriate for that search space.  As future work, we would like to incorporate dimensionality into the selection of complexity measures we present in Section~\ref{chap:complex}.

Future research will also test the ability of a single predictor to work across several domain types.  It would also be interesting to build such a predictor across search spaces of different sizes, including search spaces size as an input to the predictor.

\section{Conclusion}

In this paper, we have explored  performance of real-time heuristic search as it is affected by the properties of search spaces.  We began by formally defining heuristic search and real-time search.  We then  reviewed a selection of state of the art real-time heuristic search algorithms.  We discussed the common approach of many of these algorithms to compute a domain-specific database that provides subgoals to guide the search agent.  We explored the efforts of other researchers to understand real-time heuristic search performance, and discussed why our work was significant in this context.

We then presented our main vehicle for characterizing search spaces and understanding algorithm performance: a set of eight domain-independent complexity measures.  After explaining how the complexity measures can be efficiently computed in practice, we examined how they relate to the performance of real-time heuristic search in video-game pathfinding.  We demonstrated a statistical link between the measures and the performance of five real-time algorithms.  We then showed how the measures could be used to predict the performance of algorithms, and to assist in algorithm parameterization.  Our examination of video-game pathfinding was followed by an extension of the complexity measures to mazes and road maps.  This was the first such examination of database-driven real-time search in these domains.  Finally, we suggested avenues for future research, and discussed how this work could inform the future development of real-time heuristic search algorithms.

\section{Acknowledgements}

This work has been funded in part by the National Science and Engineering Research Council. A more detailed exposition of the results can be found in a M.Sc. dissertation~\cite{dhuntley-thesis}. A two-page extended abstract of this work was published as~\cite{huntley11}.

\small
\bibliographystyle{IEEEtran}
\bibliography{thesis}

\end{document}